\documentclass[twocolumn]{svjour3}         
\smartqed  
\usepackage{graphicx}
\usepackage{epsfig}
\usepackage{rotating}
\usepackage{subfigure}
\usepackage{amsmath}
\usepackage{amssymb}
\usepackage{natbib}

\usepackage{hyperref}
\usepackage{url}
\usepackage{array}
\usepackage{color}
\newcommand{\argmin}{\operatornamewithlimits{argmin}}
\newcommand{\minisection}[1]{\vspace{0.04in} \noindent {\bf #1}\ \ }

\begin{document}

\title{Scale Coding Bag of Deep Features for Human Attribute and Action Recognition
}


\author{Fahad~Shahbaz~Khan, Joost~van~de~Weijer, Rao~Muhammad~Anwer, Andrew D. Bagdanov, Michael Felsberg, Jorma Laaksonen
}

\date{Received: }

 \institute{Fahad Shahbaz Khan, Michael Felsberg: \at
 $^{1}$Computer Vision Laboratory, Link\"oping University, Sweden\\
 Joost van de Weijer: \at
 $^{2}$Computer Vision Centre Barcelona, Universitat Autonoma de Barcelona, Spain\\
 Rao Muhammad Anwer, Jorma Laaksonen: \at
 $^{3}$Department of Computer Science, Aalto University School of Science, Finland\\
 Andrew D. Bagdanov: \at
 $^{4}$Media Integration and Communication Center, University of Florence, Italy\\
}
%

\maketitle

\begin{abstract}
  Most approaches to human attribute and action recognition in still
  images are based on image representation in which multi-scale local
  features are pooled across scale into a single, scale-invariant
  encoding. Both in bag-of-words and the recently popular
  representations based on convolutional neural networks, local
  features are computed at multiple scales. However, these multi-scale
  convolutional features are pooled into a single scale-invariant
  representation. We argue that entirely scale-invariant
  image representations are sub-optimal and investigate approaches to
  scale coding within a Bag of Deep Features framework.

  Our approach encodes multi-scale information explicitly during
  the image encoding stage. We propose two strategies to encode
  multi-scale information explicitly in the final image
  representation. We validate our two scale coding techniques on five
  datasets: Willow, PASCAL VOC 2010, PASCAL VOC 2012, Stanford-40 and
  Human Attributes (HAT-27). On all datasets, the proposed scale
  coding approaches outperform both the scale-invariant method and the
  standard deep features of the same network. Further, combining our
  scale coding approaches with standard deep features leads to
  consistent improvement over the state-of-the-art.
\end{abstract}
\maketitle

\section{Introduction}
\label{sec:introduction}

\begin{figure*}[t!]
\centering
\subfigure[Class: interacting with computer]{
   \includegraphics[scale =0.45] {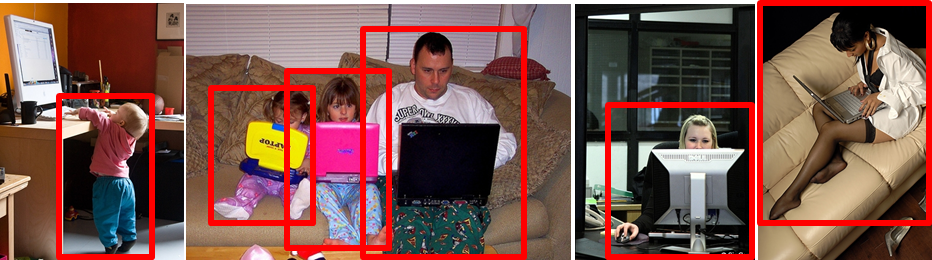}
 }
 \subfigure[Class: fishing]{
   \includegraphics[scale =0.45] {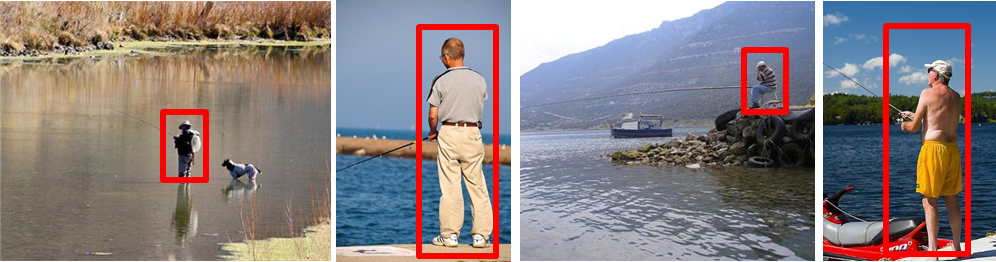}
 }
  \subfigure[Class: running]{
   \includegraphics[scale =0.45] {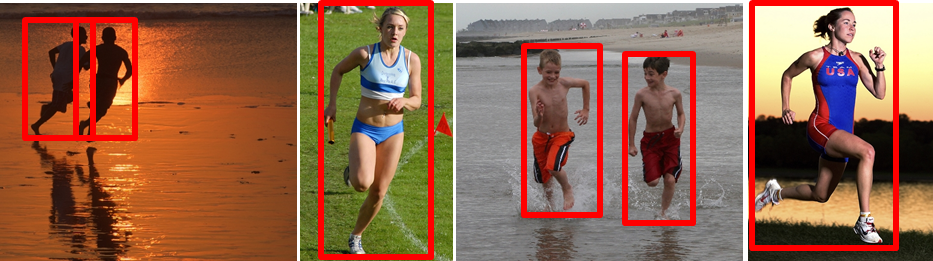}
 }
  \subfigure[Class: watching tv]{
   \includegraphics[scale =0.45] {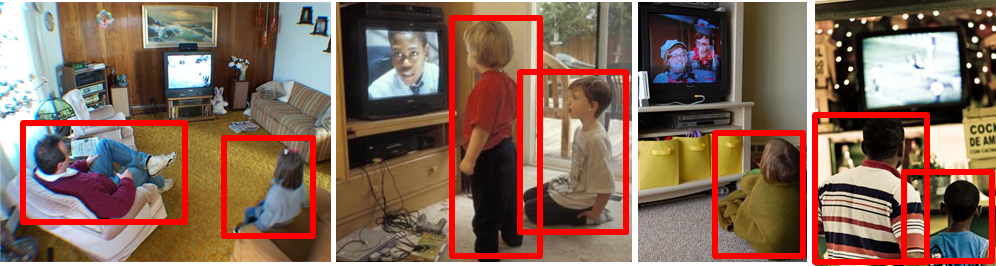}
 }
 \caption{Example images from the \emph{interacting with computer},
   \emph{fishing}, \emph{running} and \emph{watching tv} action
   categories. These examples illustrate the scale variations present,
   especially with respect to the size of bounding boxes within each
   action category. This suggests that alternative image
   representations may be desirable that explicitly encode multi-scale
   information.}
 \label{fig:intro_scale_fig}
\end{figure*}

Human attribute and action recognition in still images is a
challenging problem that has received much attention in recent
years \citep{Joo13h,fahad13b,Cordelia12c,Yao11f}. Both tasks are
challenging since humans are often occluded, can appear in different
poses (also articulated), under varying illumination, and at low resolution. Furthermore, significant
variations in scale both within and across different classes make these tasks extremely
challenging. Figure~\ref{fig:intro_scale_fig} shows example images
from different categories in the Stanford-40 and the Willow action
datasets. The bounding box (in red) of each person instance is
provided both at train and test time. These examples illustrate the
inter- and intra-class scale variations common to certain action
categories. In this paper, we investigate image representations which
are robust to these variations in scale.

Bag-of-words (BOW) image representations have been successfully
applied to image classification and action recognition
tasks \citep{fahad13b,Lazebnik06,gevers10,schmid12}. The first stage
within the framework, known as feature detection, involves detecting
keypoint locations in an image. The standard approach for feature
detection is to use dense multi-scale feature
sampling \citep{fahad12b,Jurie06,gevers10} by scanning the image at
multiple scales at fixed locations on a grid of rectangular patches.
Next, each feature is quantized against a visual vocabulary to arrive
at the final image representation.  A disadvantage of the standard BOW
pipeline is that all scale information is lost. Though for image
classification such an invariance with respect to scale might seem
beneficial since instances can appear at different scales, it trades
discriminative information for scale invariance.  We distinguish two
relevant sources of scale information: (i) dataset scale prior: due to the
acquisition of the dataset some visual words could be more indicative
of certain categories at a particular scale than at others scales
(e.g. we do not expect persons of 15 pixels nor shoes at 200 pixels)
and (ii) relative scale: in the presence of a reference scale, such as
the person bounding box provided for action recognition, we have knowledge
of the actual scale at which we expect to detect parts of the object
(e.g. the hands and head of the person). Both examples show the
relevance of scale information for discriminative image
representations, and are the motivation for our investigation into
scale coding methods for human attribute and action recognition.


Traditionally, BOW methods are based on hand-crafted local features
such as SIFT \citep{Lowe04}, HOG \citep{Dalal05} or Color
Names \citep{Weijer09a}. Recently, Convolutional Neural Networks (CNNs)
have had tremendous success on a wide range of computer vision
applications, including human attribute \citep{Zhang14h} and action
recognition \citep{Oquab14k}. Cimpoi et al. \cite{Cimpoi15c} showed how deep convolutional features
(i.e. dense local features extracted at multiple scales from the
convolutional layers of CNNs) can be exploited within a BOW pipeline
for object and texture recognition. In their approach, a Fisher Vector
encoding scheme is used to obtain the final image representation
(called FV-CNN). We will refer to this type of image representation as
a \emph{bag of deep features}, and in this work we will apply various
scale coding approaches to a bag of deep features.

\minisection{Contributions:} In this paper, we investigate strategies
for incorporating multi-scale information in image representations for
human attribute and action recognition in still images. Existing approaches encode
 multi-scale information only at the feature extraction stage by extracting convolutional features
at multiple scales. However, the final image representation in these
approaches is scale-invariant since all the scales are pooled into a
single histogram. To prevent the loss of scale information we will
investigate two complementary scale coding approaches. The first
approach, which we call \emph{absolute scale coding}, is based on a
multi-scale image representation with scale encoded with respect to
the image size. The second approach, called \emph{relative scale
  coding}, instead encodes feature scale relative to the size of the
bounding box corresponding to the person instance.  \emph{Scale coding of bag-of-deep features} is performed by applying the coding strategies to the convolutional features from a pre-trained deep network. The final image representation is obtained by concatenating the small, medium and large scale image representations. We perform comprehensive experiments on five standard datasets:
Willow, PASCAL VOC 2010, PASCAL VOC 2012, Stanford-40 and the Database
of Human Attributes (HAT-27). Our experiments clearly demonstrate that
our scale coding strategies outperform both the scale-invariant bag of
deep features and the standard deep features extracted from fully
connected layers of the same network. We further show that combining
our scale coding strategies with standard features from the FC layer
further improves the classification performance. Our scale coding
image representations are flexible and effective, while providing
consistent improvement over the state-of-the-art.

In the next section we discuss work from the literature related to our
proposed scale coding technique. In
sections~\ref{method_sec}~and~\ref{bow_deep_sec} we describe our two
proposals for coding scale information in image representations. We
report on a comprehensive set of experiments performed on five
standard benchmark datasets in section~\ref{sec:experiments}, and in
section~\ref{conclusions} we conclude with a discussion of our
contribution.

\section{Related Work}
\label{related_sec}

Scale plays an important role in feature detection. Important work includes early research on pattern spectrums \citep{Maragos89g} based on mathematical morphology which provided insight into the existence of features at certain scales in object shapes. In the field of scale-space theory \citep{Witkin84,Koenderink84g} the scale of features was examined by analyzing how images evolved when smoothed with Gaussian filters of increasing scale. This theory was also at the basis of the SIFT detector \citep{Lowe04} which obtained scale-invariant features and showed these to be highly effective for detection of objects. The detection of scale-invariant features was much studied within the context of bag-of-words \citep{Mikolajczyk04gg}. In contrast to the methods we describe in this paper, most of these works ignore the relative size of detected features.

In this section we briefly review the state-of-the-art in
bag-of-words image recognition frameworks, multi-scale deep feature
learning, and human action and attribute recognition.

\minisection{The bag-of-words framework.} In the last decade, the
bag-of-words (BOW) based image representation dominated the
state-of-the-art in object recognition \citep{everingham10b} and image
retrieval \citep{Jegou10b}. The BOW image representation is obtained by
performing three steps in succession: feature detection, feature
extraction and feature encoding.  Feature detection involves keypoint
selection either with an interest point detector \citep{Mikolajczyk04a}
or with dense sampling on a fixed
grid \citep{Bosch08g,vedaldi09h}. Several
works \citep{Rojas10j,gevers10} demonstrated the importance of using a
combination of interest point and grid-based dense sampling. This
feature detection phase, especially when done on a dense grid, is
usually \emph{multi-scale} in the sense that feature descriptors are
extracted at multiple scales at all points.

Local descriptors, such as SIFT and HOG, are extracted in the feature
extraction phase \citep{Lowe04,Dalal05}.  Next, several encoding
schemes can be considered \citep{Perronnin10c,Gemert10g,Zhou10h}. The
work of \cite{Gemert10g} investigated soft assignment of local
features to visual words. Zhou et al. \cite{Zhou10h} introduced
super-vector coding that performs a non-linear mapping of each local
feature descriptor to construct a high-dimensional sparse vector. The
Improved Fisher Vectors, introduced by Perronnin et
al. \cite{Perronnin10c}, encode local descriptors as gradients with
respect to a generative model of image formation (usually a Gaussian
mixture model (GMM) over local descriptors which serves as a visual
vocabulary for Fisher vector coding).

Regardless of the feature encoding scheme, most existing methods
achieve scale invariance by simply quantizing local descriptors to a
visual vocabulary \emph{independently} of the scale at which they were
extracted.  Visual words have no associated scale information, and scale is thus
marginalized away in the histogram construction process. In this work,
we use a Fisher Vector encoding scheme within the BOW framework and
investigate techniques to relax scale invariance in the final image
representation. We refer to this as \emph{scale coding}, since scale
information is preserved in the final encoding.

\minisection{Deep features.}  Recently, image representations based on
convolutional neural networks \citep{LeCun89h} (CNNs) have demonstrated
significant improvements over the state-of-the-art in image
classification \citep{Oquab14k}, object detection \citep{Girshick14h},
scene recognition \citep{Koskela14h}, action
recognition \citep{Liang14h}, and attribute
recognition \citep{Zhang14h}. CNNs consist of a series of convolution
and pooling operations followed by one or more fully connected (FC)
layers. Deep networks are trained using raw image pixels with a fixed
input size. These networks require large amounts of labeled training
data. The introduction of large datasets
(e.g. ImageNet \citep{ImageNet:2014}) and the parallelism enabled by
modern GPUs have facilitated the rapid deployment of deep networks for
visual recognition.

It has been shown that intermediate, hidden activations of fully
connected layers in trained deep network are general-purpose features
applicable to visual recognition tasks \citep{Azizpour14h, Oquab14k}. Several recent
methods \citep{Cimpoi15c,Gong14c,Lingqiao15c} have shown superior
performance using convolutional layer activations instead of
fully-connected ones. These convolutional layers are discriminative,
semantically meaningful and mitigate the need to use a fixed input
image size. Gong et al. \cite{Gong14c} proposed a multi-scale
orderless pooling (MOP) approach by constructing descriptors from the
fully connected (FC) layer of the network. The descriptors are
extracted from densely sampled square image windows. The descriptors
are then pooled using the VLAD encoding \citep{jegou2010aggregating}
scheme to obtain final image representation.

In contrast to MOP \citep{Gong14c}, Cimpoi et al. \cite{Cimpoi15c}
showed how deep convolutional features (i.e. dense local features
extracted at multiple scales from the convolutional layers of CNNs)
can be exploited within a BOW pipeline.  In their approach, a Fisher
Vector encoding scheme is used to obtain the final image
representation. We will refer to this type of image representation as
a \emph{bag of deep features}, and in this work we will apply various
scale coding approaches to a bag of deep features.  Though
FV-CNN \citep{Cimpoi15c} employs multi-scale convolutional features,
the descriptors are pooled into a single Fisher Vector
representation. This implies that the final image representation is
scale-invariant since all the scales are pooled into a single feature
vector. We argue that such a representation is sub-optimal for the
problem of human attribute and action recognition and propose to
explicitly incorporate multi-scale information in the final image
representation.

\minisection{Action recognition in still images.} Recognizing actions
in still images is a difficult problem that has gained a lot of
attention recently \citep{fahad13b,Oquab14k,Cordelia12c,Yao11f}. In
action recognition, bounding box information of each person instance
is provided both at train and test time. The task is to associate an
action category label with each person bounding box. Several
approaches have addressed the problem of action recognition by finding
human-object interactions in an
image \citep{Maji11f,Cordelia12c,Yao11f}. A poselet-based approach was
proposed in \cite{Maji11f} where poselet activation vectors capture
the pose of a person.  Prest et al. \cite{Cordelia12c} proposed a
human-centric approach that localizes humans and objects associated
with an action. Yao et al. \cite{Yao11f} propose to learn a set of
sparse attribute and part bases for action recognition in still
images. Recently, a comprehensive survey was performed
by \cite{Ziaeefard15c} on action recognition methods exploiting
semantic information. In their survey, it was shown that methods
exploiting semantic information yield superior performance compared
to their non-semantic counterparts in many scenarios. Human action
recognition in still images is also discussed within the context of
fuzzy domain in a recent survey \citep{Hong15c}.

Other approaches to action recognition employ BOW-based image
representations \citep{fahad13b, khan14g,schmid12}. Sharma et
al. \cite{schmid12} proposed the use of discriminative spatial
saliency for action recognition by employing a max margin
classifier. A comprehensive evaluation of color descriptors and
color-shape fusion approaches was performed by \cite{fahad13b} for
action recognition. Khan et al. \cite{khan14g} proposed pose-normalized
semantic pyramids employing pre-trained body part detectors. A
comprehensive survey was performed by \cite{Guo14c} where existing
action recognition methods are categorized based on high-level cues
and low-level features.

Recently, image representations based on deep features have achieved
superior performance for action
recognition \citep{Gkioxari14g,Hoai14k,Oquab14k}. Oquab et
al. \cite{Oquab14k} proposed mid-level image representations using
pre-trained CNNs for image classification and action recognition. The
work of \cite{Gkioxari14g} proposed learning deep features jointly for
action classification and detection. Hoai et al. \cite{Hoai14k}
proposed regularized max pooling and extract features at multiple
deformable sub-windows. The aforementioned approaches employ deep
features extracted from activations of the fully connected layers of
the deep CNNs. In contrast, we use dense local features from the
convolutional layers of networks for image description.

The incorporation of scale information has been investigated in the
context of action recognition in videos \citep{Shabani13c,Zhu13k}. The
work of \cite{Shabani13c} proposes to construct multiple dictionaries
at different resolutions in a final video representation. The work of \cite{Zhu13k} proposes multi-scale spatio-temporal concatenation of local features resulting in a set of natural action structures. Both these methods do not consider relative
scale coding. In addition, our approach is based on recent advancements of deep convolutional
neural networks (CNNs) and Fisher vector encoding scheme. We re-visit
the problem of incorporating scale information for the popular CNNs
based deep features. To the best of our knowledge, we are the first to
investigate and propose scale coded bag-of-deep feature
representations applicable for both human attribute and action
recognition in still images.

\minisection{Human attribute recognition.} Recognizing human
attributes such as, age, gender and clothing style is an active
research problem with many real-world applications. State-of-the-art
approaches employ part-based
representations \citep{Bourdev11d,khan14g,Zhang14h} to counter the
problem of pose normalization. Bourdev et al. \cite{Bourdev11d}
proposed semantic part detection using poselets and constructing
pose-normalized representations. Their approach employs HOGs for part
descriptions. Later, Zhang et al. \cite{Zhang14h} extended the
approach of \cite{Bourdev11d} by replacing the HOG features with
CNNs. Khan et al. \cite{khan14g} proposed pre-trained body part
detectors to automatically construct pose normalized semantic pyramid
representations.


In this work, we investigate scale coding strategies for human
attribute and action recognition in still images. This paper is an
extended version of our earlier work \citep{fahad14h}. Instead of using
the standard BOW framework with SIFT features, we propose scale coding
strategies within the emerging Bag of Deep Features paradigm that uses
dense convolutional features in classical BOW pipelines. We
additionally extend our experiments with results on the PASCAL VOC
2010, PASCAL 2012, Standord-40 and Human Attribute (HAT-27) datasets.

\section{Scale Coding: Relaxing Scale Invariance}
\label{method_sec}

In this section we discuss several approaches to relaxing the scale
invariance of local descriptors in the bag-of-words model.
Originally, the BOW model was developed for image classification where
the task is to determine the presence or absence of objects in
images. In such situations, invariance with respect to scale is
important since the object could be in the background of the image and
thus appear small, or instead appear in the foreground and cover most of the
image space.  Therefore, extracted features are converted to a
canonical scale --- and from that point on the original feature scale
is discarded --- and mapped onto a visual vocabulary. When BOW was
extended to object detection \citep{HJS09,vedaldi09h} and later to
action recognition \citep{Laptev10,fahad13b,Cordelia12c} this same
strategy for ensuring scale invariance was applied.

However, this invariance comes at the expense of discriminative power
through the loss of information about relative scale between
features. In particular, we distinguish two sources of scale
information: (i) dataset scale prior: the acquisition and/or collection
protocol of a data set results in a distribution of the object-sizes
as a function of the size in the image, e.g. most cars are between
100-200 pixels, and (ii) relative scale: in the presence of a
reference scale, such as the person bounding box, we have knowledge of the
actual scale at which we expect to detect parts or objects (e.g. the
size at which the action-defining object such as the mobile phone or
musical instrument should be detected). These sources of information
are lost in scale-invariant image representations. We propose two
strategies to encode scale information of features in the final image
representation.



\subsection{Scale-invariant Image Representation}
We first introduce some notation. Features are extracted from the
person bounding boxes (available at both training and testing time)
using multi-scale sampling at all feature locations.  For each
bounding box $B$, we extract a set of features:
\begin{eqnarray*}
F(B) = \left\{ \mathbf{x}_i^s \ | \ i \in \{1, \ldots, N\}, s \in \{1, \ldots M \} \right\},
\end{eqnarray*}
where $i \in \{1, \ldots, N\}$ indexes the $N$ feature sites in $B$,
and $s \in \{1, \ldots M \}$ indexes the $M$ scales extracted at each
site.


In the scale-invariant representation a single representation $h(B)$
is constructed for each bounding box $B$:
 \begin{equation}
h( B ) \propto \sum\limits_{i = 1}^N {\sum\limits_{s = 1}^M c \left( \mathbf{x}_i^s \right)}
\label{eq:scale-invariant}
\end{equation}
where $c:\Re ^p \to \Re ^q$ denotes a coding scheme which maps the
input feature space of dimensionality $p$ to the image representation
space of dimensionality $q$.

Let us first consider the case of standard bag-of-words with nearest
neighbor assignment to the closest vocabulary word. Assume we have a
visual vocabulary $W = \left\{ {\mathbf{w}_1 ,\ldots,\mathbf{w}_P } \right\}$
of $P$ words. Every feature is quantized to its closest (in the
Euclidean sense) vocabulary word:
\begin{eqnarray*}
  w_i^s = \argmin_{k \in \left\{1, \ldots, P\right\}} d(\mathbf{x}_i^s, \mathbf{w}_k),
\end{eqnarray*}
where $d(\cdot, \cdot)$ is the Euclidean distance. Index $w_i^s$
corresponds to the vocabulary word to which feature $\mathbf{x}_i^s$
is assigned. Letting $e(i)$ be the one-hot column vector of length $q$
with all zeros except for the index $i$ where it is one, we can write
the standard hard-assignment bag-of-words by plugging in:
\begin{equation*}
c_{\mathrm{BOW}}\left( {\mathbf{x}_i^s } \right) = e(w_i^s)
\end{equation*}
as the coding function in Eq.~\ref{eq:scale-invariant}.

For the case of Fisher vector encoding~\cite{perronnin2007fisher}, a
Gaussian Mixture Model (GMM) is fitted to the distribution of local
features $\mathbf{x}$:
\begin{equation*}
u_\lambda  \left( \mathbf{x} \right) = \sum\limits_1^K {w_k u_k \left( \mathbf{x} \right)},
\end{equation*}
where $\lambda = \left\{ w_k , \boldsymbol{\mu}_k ,\boldsymbol{\Sigma}_k \right\}_{k=1}^K$ are
the parameters defining the GMM, respectively the mixing weights, the
means, and covariance matrices for the $K$ Gaussian mixture
components, and
\begin{equation*}
u_k \left( \mathbf{x} \right) = \frac{1}{{\left( {2\pi } \right)^{{D \mathord{\left/
 {\vphantom {D 2}} \right.
 \kern-\nulldelimiterspace} 2}} \left| {\boldsymbol{\Sigma}_k } \right|^{{1 \mathord{\left/
 {\vphantom {1 2}} \right.
 \kern-\nulldelimiterspace} 2}} }}\exp \left\{ { - \frac{1}{2}\left( {\mathbf{x} - \boldsymbol{\mu}_k } \right)'\boldsymbol{\Sigma}_k^{-1} \left( {\mathbf{x} - \boldsymbol{\mu}_k } \right)} \right\}.
\end{equation*}
The coding function is then given by the gradient with respect to all
of the GMM parameters:
\begin{equation*}
c_{\mathrm{Fisher}}\left( {\mathbf{x}_i^s } \right) = \nabla \log u_\lambda  \left( {\mathbf{x}_i^s } \right).
\end{equation*}
and plugging this encoding function into Eq.~\ref{eq:scale-invariant}.
For more details on the Fisher vector encoding, please refer
to \cite{sanchez2013image}. Since the superiority of Fisher coding has
been shown in several publications we apply Fisher coding throughout
this paper \citep{chatfield2011devil}.


\begin{figure*}[t]
\begin{center}
\includegraphics[width=\textwidth]{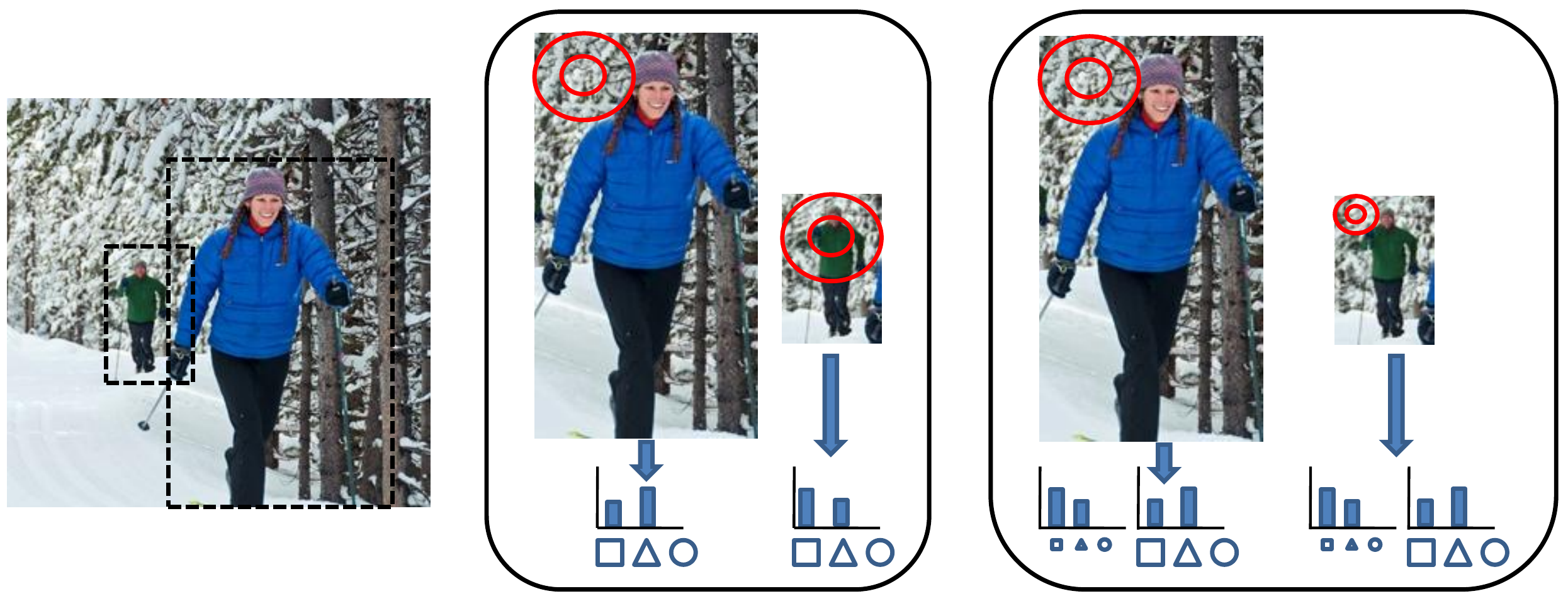}
\end{center}
\caption{Scale coding: (left) input image, superimposed bounding boxes
  indicate persons performing an action; (middle) in standard scale
  coding the scale is independent of the object size (red circles show
  the extracted feature scales), and they are all assembled in a single
  histogram per image; (right) our proposal of relative scale coding
  adapts to the bounding box of the object. This ensures that similar
  structures (such as hands and ski poles) are captured at the same
  scale independent of the absolute bounding box size. The features are
  represented in several concatenated histograms which collect a range
  of feature scales.} 
\label{fig:scale_example}
\end{figure*}


\subsection{Absolute Scale Coding}
\label{sec:absolute}
The first scale preserving scale coding method we propose uses an
absolute multi-scale image representation. Letting $S=\{1, \ldots,
M\}$ be the set of extracted feature scales, we encode features in
groups of scales:
\begin{eqnarray}
h^t (B) \propto \sum_{i=1}^N \sum_{s \in S^t } c (\mathbf{x}_i^s).
\label{MS_reprs_eq}
\end{eqnarray}
Instead of being marginalized completely away as in
equation~(\ref{eq:scale-invariant}), feature scales are instead
divided into several subgroups $S^t$ that partition the entire set of
extracted scales (i.e. $\bigcup_{t} S^t = \{1, \ldots, M\}$). In this
work we consider a split of all scales into three groups with $t \in
\left\{ {\mathrm{s},\mathrm{m},\mathrm{l}} \right\}$ for small, medium and large
scale features.  For absolute scale coding, these three scale
partitions are defined as:
\begin{eqnarray*}
S^{\rm{s}} &=& \left\{ {s \ | \ {s \le s^{\rm{s}} ,s \in S} } \right\} \\
\nonumber S^{\rm{m}} &=& \left\{ {s \ | \ {s^{\rm{s}} < s \le s^{\rm{l}} ,s \in S} } \right\} \\
\nonumber S^{\rm{l}} &=& \left\{ {s \ | \ {s^{\rm{l}} < s ,s \in S} } \right\},
\end{eqnarray*}
where the two cutoff thresholds $s^{\rm{s}}$ and $s^{\rm{l}}$ are
parameters of the encoding. The final representation is obtained by
concatenating these three encodings of the box $B$ and thus preserves
coarse scale information about the originally extracted features, and
it exploits what we refer to as the \emph{dataset scale prior} or
absolute scale. However, note that this representation does not
exploit the relative scale information.


\subsection{Relative Scale Coding}
In relative scale coding features are represented relative to the
size of the bounding box of the object (in our case the person bounding
box). The representation is computed with:
\begin{equation}
h^t(B) \propto \frac{1}{{| {\hat S^t }|}}
\sum_{i = 1}^N {\sum_{s \in \hat S^t }c(\mathbf{x}_i^s) }
\label{MS_relative_eq}
\end{equation}
The difference between Eqs.~\ref{MS_relative_eq}~and~\ref{MS_reprs_eq}
is that the scale of each feature $s$ is re-parameterized relative to the
size of the bounding box $B$ in which it was observed:
\begin{equation*}
\hat s = \frac{{B_w  + B_h }}{{\bar w + \overline h }}s
\end{equation*}
where $B_w$ and $B_h$ are the width and height of bounding box $B$ and
$\bar w$ and $\overline h$ are the mean width and hight of all
bounding boxes in the training set. Taking into account the
boundary length ensures that elongated objects have large scales.

As for absolute scale coding, described in the previous section, we
group relative scales into three groups.  The relative scale splits
$\hat{S}^t$ are defined with respect to relative scale:
\begin{eqnarray*}
{\hat S}^{\rm{s}} &=& \left\{ {\hat s \ | \ {\hat s \le s^{\rm{s}}, s \in S} }\right\} \\
\nonumber {\hat S}^{\rm{m}} &=& \left\{ {\hat s \ | \ s^{\rm{s}} < \hat s \le s^{\rm{m}}, s \in S }\right\} \\
\nonumber {\hat S}^{\rm{l}} &=& \left\{ {\hat s \ | \ {s^{\rm{m}} < \hat s, s \in S} }\right\}.
\end{eqnarray*}
Since the number of scales which fall into the small, medium and large
scale range image representation now varies with the size of the
bounding box, we introduce a normalization factor ${| {{\hat S}^t }|}$
in Eq.~\ref{MS_relative_eq} to counter this. Here ${| {{\hat S}^t
  } |}$ is the cardinality of the set ${\hat S}^t$.

Relative scale coding represents visual words at a certain relative
scale with respect to the bounding box size. Again, it consists of
three image representations for small, medium and large scale visual
words, which are then concatenated together to form the final
representation for $B$. However, depending on the size of the bounding
box, the scales which are considered small, medium and large
change. An illustrative overview of this approach is given in
Figure~\ref{fig:scale_example}. In contrast to the standard approach,
this method preserves the relative scale of visual words without
completely sacrificing the scale invariance of the original
representation.


\subsection{Scale Partitioning}
Until now we we have considered partitioning the features into three
scale-groups: small, medium and large. Here, we evaluate this choice
and compare it with other partitioning of the scales.

\begin{figure}[t]
\centerline{\includegraphics[width=8.2cm]{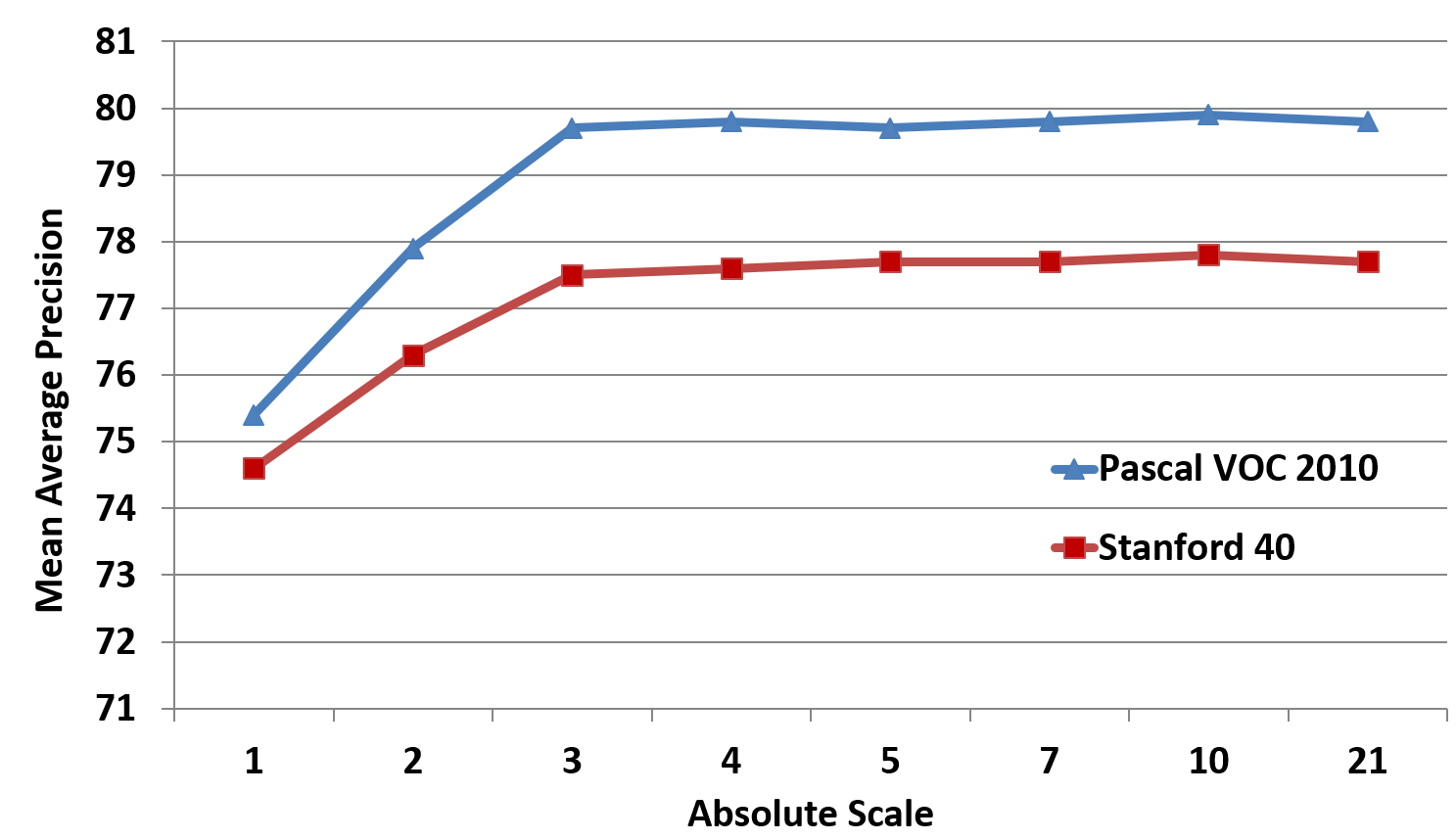}}
\caption{Mean average precision as a function of number of scale
  partitions for absolute scale coding
  (section~\ref{sec:absolute}). Performance is shown on the Stanford-40
  and PASCAL VOC 2010 validation sets. On both, absolute scale
  coding improves performance compared to scale-invariant coding
  (which groups all scales to single representation). A consistent
  improvement is achieved when using three scale partitions for
  absolute coding. }
\label{fig:scale-partitioning}
\end{figure}

To evaluate the partitioning of scales, we extracted features at
$M=21$ different scales on Stanford-40 and the PASCAL VOC 2010
datasets. For this evaluation we performed absolute scale encoding and
varied the number of scale partitions from one (equivalent to standard
scale-invariant coding) to 21 in which case every scale is represented
by a single image representation. In
Figure~\ref{fig:scale-partitioning} we plot the mean average precision
(mAP) on Stanford-40 and PASCAL 2010 as a function of the number of
scale partitions. The curve clearly shows that absolute scale coding
outperforms the generally applied representation based on
scale-invariant coding (which collects all scales in a single
partition). Furthermore, it shows that after three scale partitions,
the gain of increasing the number of partitions is
negligible. Throughout this paper we use three scale partitions for
all scale coding experiments.

\section{The Bag of Deep Features Model}
\label{bow_deep_sec}


Inspired by the recent success of CNNs, we use deep features in our
scale coding framework.

\minisection{Deep convolutional features:} Similar
to \cite{Cimpoi15c}, we use the VGG-19 network proposed by \cite{Simonyan15j}, pre-trained on the ImageNet dataset. It was shown to provide the best performance in a recent
evaluation \citep{Cimpoi15c,Chatfield14h} for image classification
tasks. In the VGG-19 network, input images are convolved with $3
\times 3$ filters at each pixel at a stride of 1 pixel. The network
contains several max-pooling layers which perform spatial pooling over
$2\times2$ pixel windows at a stride of 2 pixels. The VGG-19 network
contains 3 fully connected (FC) layers at the end. The width of the
VGG-19 network starts from 64 feature maps in the first layer and
increases by a factor of 2 after each max-pooling layer to reach 512
feature maps at its widest (see \citep{Simonyan15j} for more details).

Typically, the activations from the FC layer(s) are used as input
features for classification. For VGG-19 this results in a
4096-dimensional representation. In contrast, we use the output of the
last convolutional layer of the network since it was shown to provide
superior performance compared to other layers \citep{Cimpoi15c}. This
layer returns dense convolutional features at a stride of eight
pixels. We use these 512-dimensional descriptors as local features
within our scale coding framework. To obtain multi-scale samples, we
rescale all images over a range of scales and pass them through the
network for feature extraction. Note that the number of extracted
local convolutional patches depend on the size of the input image.

\minisection{Vocabulary construction and assignment.} In standard BOW
all features are quantized against a scale-invariant visual
vocabulary. The local features are then pooled in a single
scale-invariant image representation. Similar to \cite{Cimpoi15c},
we use the Fisher Vector encoding for our scale coding models. For
vocabulary construction, we use the Gaussian mixture model (GMM). The convolutional features are then pooled via the Fisher encoding that captures the average first and  second order differences. The 21 different scales are pooled into the three scale partitions to ensure that the scale information is preserved in the final representation. It is worth
mentioning that our scale coding schemes can also be used with other
encoding schemes such as hard assignment, soft assignment, and
VLAD \citep{jegou2010aggregating}.

\section{Experimental Results}
\label{sec:experiments}
In this section we present the results of our scale coding strategies
for the problem of human attribute and action recognition. First we
detail our experimental setup and datasets used in our evaluation, and
then present a comprehensive comparison of our approach with baseline
methods. Finally, we compare our approach with the state-of-the-art in
human attribute and action recognition.

\subsection{Experimental Setup}
\label{sec:experiment_setup}
As mentioned earlier, bounding boxes of person instances are provided
at both train and test time in human attribute and action recognition.
Thus the task is to predict the human attribute or action category for
each person bounding box. To incorporate context information, we
extend each person bounding box by $50\%$ of its width and height.

In our experiments we use the pre-trained VGG-19
network \citep{Simonyan15j}. Similar to Cimpoi et al. \cite{Cimpoi15c},
we extract the convolutional features from the output of the last
convolutional layer of the VGG-19 network. The convolutional features are not de-correlated by using PCA before employing Fisher Vector encoding, since it has been shown \citep{Cimpoi15c} to deteriorate the results. The convolutional features
are extracted after rescaling the image at 21 different scales $s \in
\{0.5 + 0.1 n \ | \ n = 0, 1, \ldots, 20 \}$. This results in
512-dimensional dense local features for each scaled image. On an image of size $300 \times 300$, the feature extraction on multi-core CPU takes about 5 seconds. For our scale coding approaches, we keep a single, constant threshold for all datasets.

\begin{figure}[t]
\centerline{\includegraphics[width=8.2cm]{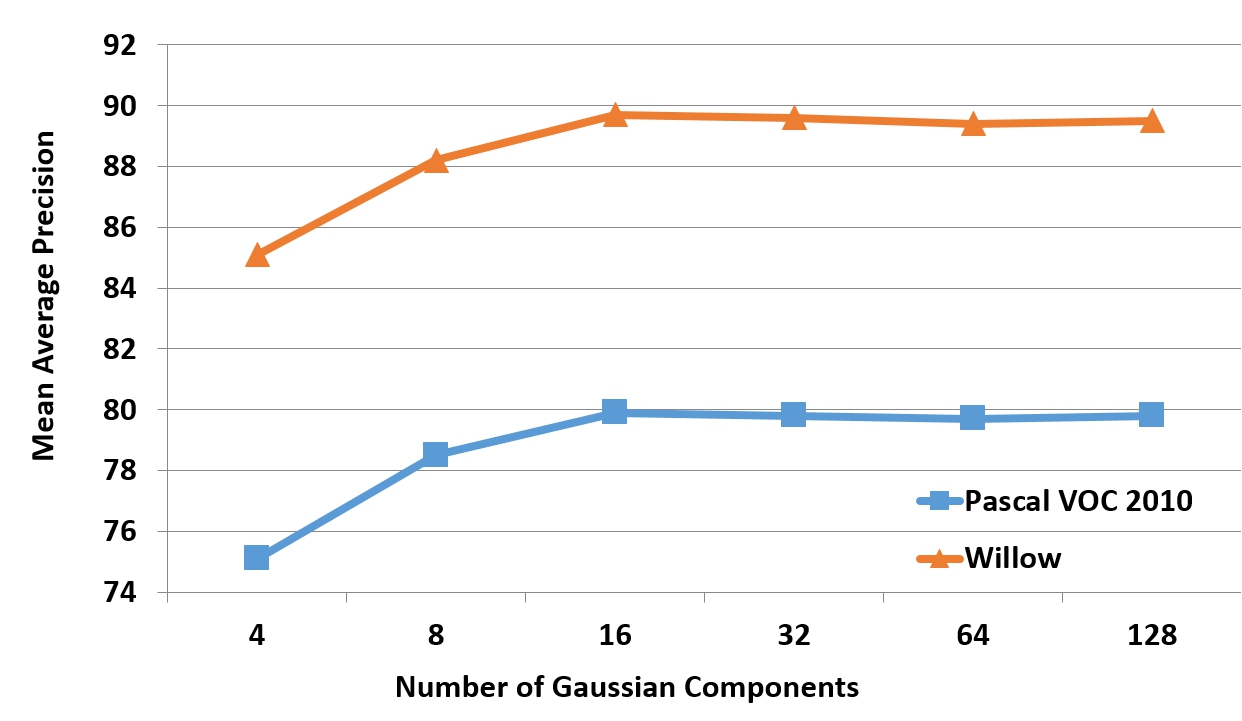}}
\caption{Mean average precision as a function of number of Gaussian components for relative scale coding. Performance is shown on the Willow
  and PASCAL VOC 2010 sets. On both, our scale
  coding provides best results when using a
Gaussian Mixture Model (GMM) with 16 components. }
\label{fig:gaussian-components}
\end{figure}

For each problem instance, we construct a visual vocabulary using a
Gaussian Mixture Model (GMM) with 16 components. In
Figure~\ref{fig:gaussian-components} we plot the mean average precision
(mAP) on Willow and PASCAL 2010 datasets as a function of the number of Gaussian components. We observed no
significant gain in classification performance by increasing the
number of Gaussian components beyond 16. The parameters of this model
are fit using a set of dense descriptors sampled from descriptors over
all scales on the training set. We randomly sample 100 descriptor
points from each training image. The resulting sampled feature
descriptors from the whole training set are then used to construct a
GMM based dictionary. We also perform experiments by varying the
number of feature samples per image. However, no improvement in
performance was observed with increased feature samples per image. We
employ a GMM with diagonal covariances. Finally, the Fisher vector
representations discussed in section~\ref{method_sec} are constructed
for each image. The Fisher vector encoding is performed with respect
to the Gaussian mixture model (GMM) with means, diagonal covariances,
and prior probabilities. The dense image features have dimensionality
512, and so our final scale-coded Fisher vector representation has
dimensionality $3 \times (2 \times 16 \times 512 + 16) = 49200$ (i.e. it is a
concatenation of the Fisher vector encoding the three scale
categories).  In our experiments, we use the standard VLFeat
library \citep{Vedaldi10b} commonly used to construct GMM based
vocabulary and the improved Fisher vector based image
representations. For classification, we employ SVMs with linear
kernels on the concatenated Fisher vectors of each scale-coding groups
described above.

\vspace{-0.8cm}



\begin{table*}
\centering
\scriptsize\addtolength{\tabcolsep}{-1.75pt}
\begin{tabular}{|r||c|c|c|c|c|c||c|}
\hline  & \textbf{Willow} & \textbf{PASCAL 2010} & \textbf{PASCAL 2012}  & \textbf{Stanford-40} & \textbf{HAT-27}  \\\hline
\hline VGG-19 FC \cite{Simonyan15j}   & $87.1$ & 72.0 & $74.0$ & $70.3$ & 61.2 \\
\hline MOP \cite{Gong14c}   & $87.6$ & 74.8 & $75.3$ & $74.2$ & 64.1  \\
\hline FV-CNN \cite{Cimpoi15c}   & $87.9$ & 75.4 & $75.6$ & $74.6$ & 64.5  \\
\hline FV-CNN-SP    & $88.4$ & 78.1 & $77.3$ & $76.9$ & 66.6  \\
\hline

\hline
\hline  Absolute-Scale Coding & $89.3$ & 79.7 & $78.1$ & $77.5$ &67.3  \\
\hline  Relative-Scale Coding & $89.7$ & 79.9 & $78.4$ & $77.8$ & 67.4  \\
\hline   Absolute + Relative + FC& $\textbf{92.1}$ & \textbf{82.7} & $\textbf{80.3}$ & $\textbf{80.0}$ & \textbf{70.6}  \\
\hline
\end{tabular}

\caption{%
  Comparison (in mAP) of the standard deep features (FC, for ``fully connected''),
  the MOP approach, the baseline scale-invariant approach (FV-CNN), the scale-invariant spatial pyramid approach (FV-CNN-SP), and our proposed relative and absolute scale coding schemes. Scale coding yields consistent improvements on all 5 datasets.
}

\label{baseline_comp_tab}
\end{table*}

\subsection{Datasets}
\label{sec:dataset}
We perform experiments on five datasets to validate our approach:
\begin{itemize}
\item The \textbf{Willow Action Dataset} consisting of seven action categories:
\emph{interacting with computer, photographing, playing music, riding
  bike, riding horse, running and walking}.\footnote{Willow
  is available at:
  \url{http://www.di.ens.fr/willow/research/stillactions/}}
\item The \textbf{Stanford-40 Action Dataset} consisting of $9532$
  images of $40$ different action categories such as \emph{gardening,
    fishing, applauding, cooking, brushing teeth, cutting vegetables,
    and drinking}.\footnote{Stanford-40 is at
    \url{http://vision.stanford.edu/Datasets/40actions.html}}
\item The \textbf{PASCAL VOC 2010 Action Dataset} consisting of $9$ action
  categories: \emph{phoning, playing instrument, reading, riding bike,
    riding horse, running, taking photo, using computer and
    walking}.\footnote{PASCAL 2010 is at:
    \url{http://www.pascal-network.org/challenges/VOC/voc2010/}}
\item The \textbf{PASCAL VOC 2012 Action Dataset} consisting of $10$ different
  action classes: \emph{phoning, playing instrument, reading, riding
    bike, riding horse, running, taking photo, using computer, walking
    and jumping}.\footnote{PASCAL 2012 is at:
    \url{http://www.pascal-network.org/challenges/VOC/voc2012/}}.
\item The \textbf{27 Human Attributes Dataset (HAT-27)} consisting of
  9344 images of 27 different human attributes such as
  \emph{crouching, casual jacket, wedding dress, young and
    female}.\footnote{HAT-27 is available at:
    \url{https://sharma.users.greyc.fr/hatdb/}}.
\end{itemize}

The test sets for both the PASCAL VOC 2010 and 2012 datasets are
withheld by the organizers and results must be submitted to an
evaluation server. We report the results on the test sets in
section~\ref{sec:soa_comp_sub} and provide a comparison with
state-of-the-art methods. For the Willow \citep{Laptev10},
Stanford-40 \citep{Yao11f} and HAT-27 \citep{SharmaHAT27:2011} datasets
we use the train and test splits provided by the respective authors.

\minisection{Evaluation criteria:} We follow the same evaluation
protocol as used for each dataset. Performance is measured in average
precision as area under the precision-recall curve. The final
performance is calculated by taking the mean average precision (mAP)
over all categories in each dataset.

%
%
%

\begin{table*}[t]
\centering
\resizebox{\textwidth}{!} {
\begin{tabular}{|r||ccccccc||c|}
  \hline
    & \textbf{int. computer} & \textbf{photographing} & \textbf{playingmusic} & \textbf{ridingbike} & \textbf{ridinghorse} & \textbf{running} & \textbf{walking} & \textbf{mAP} \\ \hline \hline

  BOW-DPM  \cite{Laptev10} & 58.2 & 35.4 & 73.2 & 82.4 & 69.6 & 44.5 & 54.2 & 59.6  \\
  POI  \cite{Laptev11j} & 56.6 & 37.5 & 72.0 & 90.4 & 75.0 & 59.7 & 57.6 & 64.1  \\
  DS  \cite{schmid12} & 59.7 & 42.6 & 74.6 & 87.8 & 84.2 & 56.1 & 56.5 & 65.9  \\
  CF  \cite{fahad13b} & 61.9 & 48.2 & 76.5 & 90.3 & 84.3 & 64.7 & 64.6 & 70.1  \\
  EPM  \cite{sharma13h} & 64.5 & 40.9 & 75.0 & 91.0 & 87.6 & 55.0 & 59.2 & 67.6  \\
  SC  \cite{fahad14h} & 67.2 & 43.9 & 76.1 & 87.2 & 77.2 & 63.7 & 60.6 & 68.0  \\
  SM-SP  \cite{khan14g} & 66.8 & 48.0 & 77.5 & 93.8 & 87.9 & 67.2 & 63.3 & 72.1  \\
  EDM  \cite{Liang14h} & 86.6 & \textbf{90.5} & 89.9 & 98.2 & 92.7 & 46.2 & 58.9 & 80.4  \\
  NSP  \cite{Mettes15g} & 88.6 & 61.8 & 93.4 & 98.8 & 98.4 & 69.4 & 62.3 & 81.7  \\
  DPM-VR  \cite{Sicre15c} & 84.9 & 72.0 & 91.2 & 96.9 & 93.6 & 73.4 & 61.0 & 81.9  \\
   \hline\hline
  \textbf{This Paper} & \textbf{96.6} & 89.2 & \textbf{98.2} & \textbf{99.8} & \textbf{99.3} & \textbf{83.0} & \textbf{78.7} & \textbf{92.1}  \\
  \hline

  \hline
\end{tabular}
}
\caption{Comparison of our approach with the state-of-the-art
  on the Willow dataset. Our proposed approach achieves best results on 6 out of 7 action categories.} 
\label{willowsoacomp_tab}
\end{table*}

\begin{table*}[t]
\centering
\resizebox{\textwidth}{!} {
\begin{tabular}{|r||ccccccccc||c|}
  \hline

    & \textbf{phoning} & \textbf{playingmusic} & \textbf{reading} & \textbf{ridingbike} & \textbf{ridinghorse} & \textbf{running} & \textbf{takingphoto} & \textbf{usingcomputer} & \textbf{walking} & \textbf{mAP} \\ \hline \hline
  Poselets \cite{Maji11f} & 49.6 & 43.2 & 27.7 & 83.7 & 89.4 & 85.6 & 31.0 & 59.1 & 67.9 & 59.7  \\
  IaC  \cite{Pedersoli11} & 45.5 & 54.5 & 31.7 & 75.2 & 88.1 & 76.9 & 32.9 & 64.1 & 62.0 & 59.0  \\
  POI  \cite{Laptev11j} & 48.6 & 53.1 & 28.6 & 80.1 & 90.7 &  85.8 & 33.5 & 56.1 & 69.6 & 60.7  \\
  LAP  \cite{Yao11f} & 42.8 & 60.8 & 41.5 & 80.2 & 90.6 &  87.8 & 41.4 & 66.1 & 74.4 & 65.1  \\
 WPOI  \cite{Cordelia12c} & 55.0 & 81.0 & 69.0 & 71.0 & 90.0 &  59.0 & 36.0 & 50.0 & 44.0 & 62.0  \\

  CF \cite{fahad13b} & 52.1 & 52.0 & 34.1 & 81.5 & 90.3 &  88.1 & 37.3 & 59.9 & 66.5 & 62.4  \\
  SM-SP  \cite{khan14g} & 52.2 & 55.3 & 35.4 & 81.4 & 91.2 &  89.3 & 38.6 & 59.6 & 68.7 & 63.5  \\
\hline\hline
\textbf{This Paper} & \textbf{64.3} & \textbf{94.5} & \textbf{65.1} & \textbf{96.9} & \textbf{96.8} &  \textbf{93.4} & \textbf{77.1} & \textbf{87.7} & \textbf{78.9} & \textbf{83.7}  \\
  \hline
\end{tabular}
}
\caption{Comparison with the state-of-the-art results on the PASCAL VOC 2010 \emph{test} set. Our scale coding based approach provides consistent improvements compared to existing methods. }
\label{pascalsoacomp2010_tab}
\end{table*}

\begin{table*}[t]
\resizebox{\textwidth}{!} {
\begin{tabular}{|r||cccccccccc||c|}
  \hline
    & \textbf{phoning} & \textbf{playingmusic} & \textbf{reading} &
    \textbf{ridingbike} & \textbf{ridinghorse} & \textbf{running} &
    \textbf{takingphoto} & \textbf{usingcomputer} & \textbf{walking} &
    \textbf{jumping} & \textbf{mAP} \\ \hline \hline
  Stanford  & 44.8 & 66.6 &44.4 & 93.2 & 94.2 & 87.6 & 38.4 & 70.6 & 75.6 & 75.7 &69.1  \\
  Oxford  & 50.0 & 65.3 &39.5 & 94.1 & 95.9 & 87.7 & 42.7 & 68.6 & 74.5 & 77.0 &69.5  \\
  Action Poselets \cite{Maji11f}  & 32.4 & 45.4 & 27.5 & 84.5 & 88.3 & 77.2 & 31.2 & 47.4 & 58.2 & 59.3 &55.1  \\
  MDF \cite{Oquab14k}  & 46.0 & 75.6 &45.3 & 93.5 & 95.0 & 86.5 & 49.3 & 66.7 & 69.5 & 78.4 &70.2  \\
  WAB \cite{Hoai14h}  & 49.5 & 67.5 & 39.1 & 94.3 & 96.0 & 89.2 & 44.5 & 69.0 & \textbf{75.9} & 79.6 &70.5  \\
  Action R-CNN \cite{Gkioxari14g}  & 47.4 & 77.5 & 42.2 & 94.9 & 94.3 & 87.0 & 52.9 & 66.5 & 66.5 & 76.2 &70.5  \\
RMP \cite{Hoai14k}  & 52.9 & 84.3 & 53.6 & 95.6 & 96.1 & 89.7 & 60.4 & 76.0 & 72.9 & 82.3 &76.4  \\
TL \cite{Khan15k}  & 62.4 & 91.3 & 61.1 & 93.3 & 95.1 & 84.1 & 59.8 & 84.5 & 53.0 & 84.9 &77.0  \\
VGG-19 + VGG-16 &  & & & & & & & & & &  \\
+ Full Image \cite{Simonyan15j}  & \textbf{71.3} & \textbf{94.7} &\textbf{71.3} & 97.1 & \textbf{98.2} & \textbf{90.2} & 73.3 & \textbf{88.5} & 66.4 & 89.3 &\textbf{84.0}  \\
  \hline
  \hline

\textbf{This Paper} & 69.7 & 92.4 & 70.8 & \textbf{97.2} & 98.0 &  89.8 & \textbf{73.8} & 88.4 & 69.4 & \textbf{89.5} & 83.9  \\

  \hline
\end{tabular}
}
\caption{Comparison of our proposed approach with the state-of-the-art on the
  PASCAL VOC 2012 \emph{test} set. The best existing results are obtained by
  combining FC features from two CNNs (VGG-16 and VGG-19). The
  features are extracted both from the full-image and  bounding box of a person.
  Our approach, based only on VGG-19 network and without using
  full-image information, obtains comparable performance with best
  results on 3 out of 10 action categories.}
\label{table:pascal2012_classification}
\end{table*}

\subsubsection{Baseline Scale-Coding Performance Analysis}
We first give a comparison of our scale coding strategies with the
baseline scale-invariant coding. Our baseline is the FV-CNN
approach \citep{Cimpoi15c} where multi-scale convolutional features are
pooled in a single scale-invariant image representation. The FV-CNN approach is further extended with spatial information by employing spatial pyramid pooling scheme \citep{Lazebnik06}. The spatial pyramid scheme is used with two levels ($1\times1$ and $2\times2$), yielding a total of 5 cells. We also compare our results with standard deep features obtained from the
activations of the first fully connected layer of the
CNN. Additionally, we compare our approach with Multiscale Orderless
Pooling (MOP) \citep{Gong14c} by extracting FC activations at three
levels: 4096-dimensional CNN activation from the entire image patch
(the person bounding box), $128 \times 128$ patches of 4096 dimensions
pooled using VLAD encoding with 100 visual-words, and the same VLAD
encoding but with $64 \times 64$ patches. The three representations
are concatenated into a single feature vector for classification. Note
that we use the same VGG-19 network for all of these image encodings.

Table~\ref{baseline_comp_tab} gives the baseline comparison on all
five datasets. Since the PASCAL VOC 2010 and 2012 test sets are
withheld by the organizers, performance is measured on the validation
sets for the baseline comparison. The standard multi-scale invariant
approach (FV-CNN) improves the classification performance compared to
the standard FC deep features. The spatial pyramid based FV-CNN further improves over the standard FV-CNN method. Our absolute and relative scale coding
approaches provide a consistent gain in performance on all datasets, compared to
baselines using features from the same deep network. Note
that the standard scale-invariant (FV-CNN) and our scale coding
schemes are constructed using the same visual vocabulary (GMM) and set
of local features from the convolutional layer. Finally, a further
gain in accuracy is obtained by combining the classification scores of
our two scale coding approaches with the standard FC deep
features. This combination is done by simply adding the three
classifier outputs. On the Stanford-40 and HAT-27 datasets, this
approach yields a considerable gain of $6.5\%$ and $4.8\%$ in mAP,
respectively, compared to the MOP approach employing FC features from
the same network (VGG-19). These results suggest that the FC, absolute
scale, and relative scale encodings have complementary information
that when combined yield results superior to each individual
representation.

\begin{figure}[t]
\begin{center}

\includegraphics[width=10.2cm,trim=20 228 80 280,clip=true]{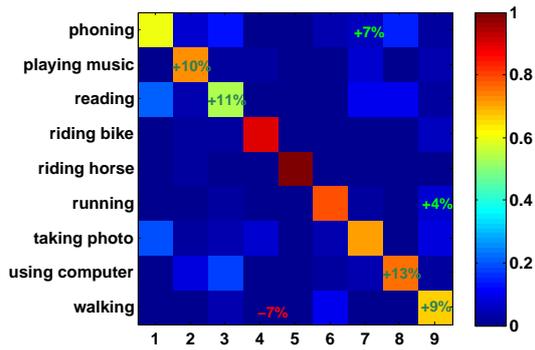}

\end{center}\vspace{-1.15cm}
\caption{Confusion matrix for our approach, combining both absolute
  and relative scale coding, on PASCAL VOC 2010. We superimposed the
  differences with the confusion matrix for the scale-invariant FV-CNN
  approach for confusions where the absolute change is at least
  $4\%$. Our approach provides consistent improvements, in general,
  but improves significantly the performance for playing music
  ($10\%$), reading ($11\%$) and using computer ($13\%$) categories.
}
\label{fig:CF_Pascal}
\end{figure}

\subsubsection{Comparison with the State-of-the-art}
\label{sec:soa_comp_sub}

We now compare our approach with the state-of-the-art on the five
benchmark datasets. In this section we report results for the
combination of our relative and absolute scale coding strategies with
the FC deep features. The combination is done by simply adding the
three classifier outputs.

\minisection{Willow:} Table~\ref{willowsoacomp_tab} gives a comparison
of our combined scale coding approach with the state-of-the-art on the
Willow dataset. Our approach achieves the best performance reported on
this dataset, with an mAP of $92.1\%$. The shared part detectors
approach of \cite{Mettes15g} achieves an mAP of $81.7\%$, while the
part-based deep representation approach \citep{Sicre15c} obtains an mAP
of $81.9\%$. Our approach, without exploiting any part information,
yields the best results on 6 out of 7 action categories, with an
overall gain of $10.2\%$ in mAP compared to \cite{Sicre15c}.

\minisection{PASCAL VOC 2010:} Table~\ref{pascalsoacomp2010_tab}
compares our combined scale coding approach with the state-of-the-art
on the PASCAL VOC 2010 Action Recognition \emph{test} set. The color
fusion approach of \cite{fahad13b} achieves an mAP of $62.4\%$, the
semantic pyramid approach by Khan et al. \cite{khan14g} obtains a mAP
of $63.5\%$, and the method of \cite{Yao11f} based on learning a
sparse basis of attributes and parts achieves an mAP of $65.1\%$. Our
approach yields consistent improvement over the state-of-the-art with
an mAP of $83.7\%$ on this dataset. Figure~\ref{fig:CF_Pascal} shows
the confusion matrix for our scale coding based approach. The
differences with the confusion matrix based on the standard
scale-invariant FV-CNN approach are superimposed for confusions where
the absolute change is at least $4\%$. Overall, our approach improves
the classification results with notable improvements for playing music
($10\%$), reading ($11\%$) and using computer ($13\%$) action
categories. Further, our approach reduces confusion all categories
except for walking.


\begin{table}[t]
\centering \tiny\addtolength{\tabcolsep}{0.0000001pt}

\begin{tabular}{|r|c|c|c|c|c|c|c|}
\hline  &  \textbf{SB}  & \textbf{CF}  & \textbf{SM-SP}  &  \textbf{Place}  & \textbf{D-EPM} &  \textbf{TL}  & \textbf{Ours} \\
\hline\hline \textbf{mAP}   &  45.7 & $51.9$ & $53.0$ & 55.3 & 72.3 & 75.4  & \bfseries 80.0 \\

\hline

\end{tabular}

\caption{
  Comparison of the proposed approach with the state-of-the-art methods on Stanford-40 dataset. Our approach yields a
  significant gain over the best reported
  results in the literature.
}
\label{SOA_cls_stanford_det_tab}
\end{table}

\minisection{PASCAL VOC 2012:} In
Table~\ref{table:pascal2012_classification} we compare our approach
with state-of-the-art on the PASCAL VOC 2012 Action Recognition
\emph{test} set. Among existing approaches, Regularized Max Pooling
(RMP) \citep{Hoai14k} obtains a mAP score of $76.4\%$.  The best
results on this dataset are obtained by combining the FC features of
the VGG-16 and VGG-19 networks. These FC features are extracted both
from the full-image and the provided bounding box of the person.  Our
combined scale coding based approach provides the best results on 3
out of 10 action categories, and achieves an mAP of $83.9\%$ on the
PASCAL 2012 test set. It is worth mentioning that our scale coding
based approach employs a single network (VGG-19) and does not exploit
the full-image information. Combining our Scale coding based
approaches using multiple deep networks is expected to further improve
performance.


\minisection{Stanford-40 dataset:} In
Table~\ref{SOA_cls_stanford_det_tab} we compare scale coding with
state-of-the-art approaches: SB \cite{Yao11f}, CF \cite{fahad13b}, SM-SP \cite{khan14g},  Place \cite{Zhou14k}, D-EPM \cite{Sharma15g} and TL \cite{Khan15k}.
Stanford-40 is the most challenging action dataset and contains 40
categories. The semantic pyramids of Khan et al. \cite{khan14g}
achieve an mAP of $53.0\%$. Their approach combines spatial pyramid
representations of full-body, upper-body and face regions using
multiple visual cues. The work of \cite{Zhou14k} uses deep features
trained on ImageNet and a recently introduced large scale dataset of
Place Scenes. Their hybrid deep features based approach achieves a mAP
of $55.3\%$. The D-EPM approach \citep{Sharma15g} based on expanded
part models and deep features achieves a mAP score of $72.3\%$. The
transfer learning (TL) based approach \citep{Khan15k} with deep
features obtains a mAP score of $75.4\%$. Our combined scale coding
approach achieves state-of-the-art results with a gain of $4.6\%$ in
mAP compared to the TL based approach \citep{Khan15k}.


\begin{figure*}[ht]
\begin{center}
\includegraphics[width=\textwidth,trim=20 48 10 68,clip=true]{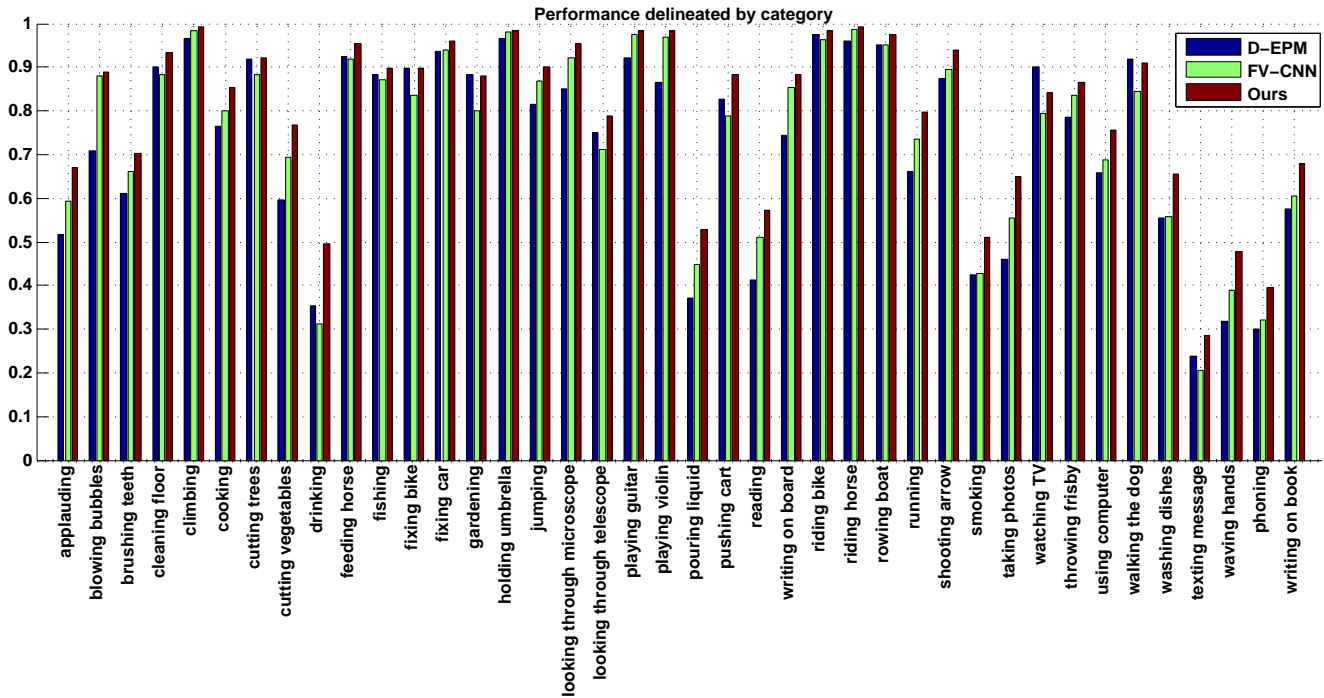}

\end{center}\vspace{-1.95cm}
\caption{Per-category performance comparison (in AP) of our approach
  with the D-EPM method \citep{Sharma15g} and the scale-invariant
  FV-CNN approach \citep{Cimpoi15c}. Our approach improves the results
  on 37 out 40 action classes compared to these two methods.}
\label{fig:per_class_stanford}
\end{figure*}

\begin{center}
\begin{table*}[!ht]
\resizebox{\textwidth}{!}{%
\begin{tabular}{||p{3cm}||p{3cm}||p{3cm}||p{3cm}||p{3cm}||}

\multicolumn{5}{c}{  \bfseries Ranking of Different Action Categories}\\[5pt] \hline\hline
 \bfseries Method  & \includegraphics[width=3cm,height=3cm]{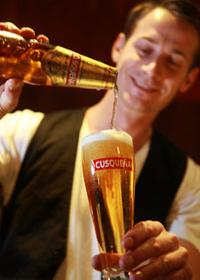}  &
 \includegraphics[width=3cm,height=3cm]{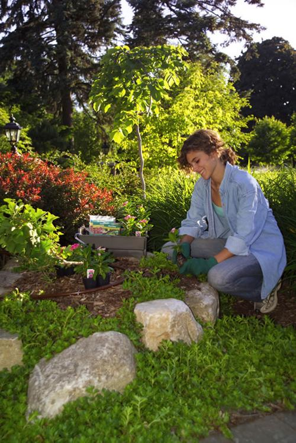}&
 \includegraphics[width=3cm,height=3cm]{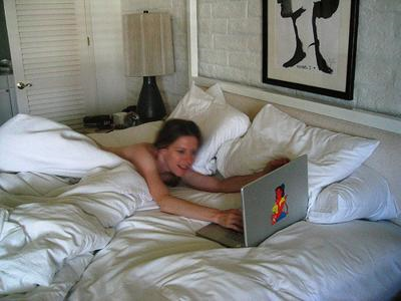} &
 \includegraphics[width=3cm,height=3cm]{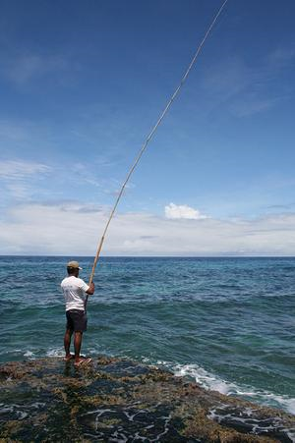}

 \\
\hline\hline
 \bfseries VGG-19 FC \cite{Simonyan15j} &186 (98) &51 (5) &104 (24) & 56 (5) \\ \hline
 \bfseries FV-CNN \cite{Cimpoi15c} &144 (62) &57 (4) & 98 (22) & 52 (5) \\ \hline
 \bfseries Our Approach &   32 (5) & 17 (1) &30 (1) &  38 (2) \\ \hline
\end{tabular}
}
\caption{
  Images from pouring liquid, gardening, using computer and fishing
  action categories from the Stanford-40 data set. The number indicates
  the absolute rank of corresponding image in the list of all test
  images sorted by the probability for the corresponding class. The
  number in parentheses after each rank is the number of false
  positives appearing \emph{before} the example test image in the
  ranked list. Lower absolute rank reflects
  higher confidence in the class label. The action category list
  contains 5532 test instances. Our approach outperforms both VGG-19 FC
  and FV-CNN methods on these images demonstrating the importance of
  coding multiple scales in the final image representation.
}
\label{stanford_rank_tab}
\end{table*}
\end{center}

%
\begin{table*}[t]
\resizebox{\textwidth}{!} {
\begin{tabular}{|r|cccccccccccccc|}
  \hline
    & \textbf{female} & \textbf{frontalpose} & \textbf{profilepose} & \textbf{turnedback} & \textbf{upperbody} & \textbf{standing} & \textbf{runwalk} & \textbf{crouching} & \textbf{sitting} & \textbf{armsbent} & \textbf{elderly} & \textbf{middleaged} & \textbf{young} & \textbf{teen}  \\ \hline \hline
  \textbf{EPM} \cite{sharma13h} & 85.9 & 93.6 &67.3 & 77.2 & 97.9 & 98.0 & 74.6 & 24.0 & 62.7 & 94.0 & 38.9 & 68.9 & 64.2 & 36.2 \\
  \textbf{RAD} \cite{Joo13h} & 91.4 & \textbf{96.8} &\textbf{77.2} & \textbf{89.8} & 96.3 & 97.7 & 63.5 & 12.3 & 59.3 & 95.4 & 32.1 & 70.0 & 65.6 & 33.5 \\
\textbf{SM-SP} \cite{khan14g} & 86.1 & 92.2 &60.5 & 64.8 & 94.0 & 96.6 & 76.8 & 23.2 & 63.7 & 92.8 & 37.7 & 69.4 & 67.7 & 36.4 \\
\textbf{D-EPM} \cite{Sharma15g} & \textbf{93.2} & 95.2 &72.6 & 84.0 & \textbf{99.0} & 98.7 & 75.1 & \textbf{34.2} & 77.8 & 95.4 & 46.4 & 72.7 & 70.1 & 36.8 \\
\hline\hline
\textbf{This Paper} & 92.0 & 95.7 &62.9 & 86.9 & 95.1 & \textbf{98.8} & \textbf{80.3} & 31.6 & \textbf{87.0} & \textbf{95.5} & \textbf{54.7} & \textbf{74.6} & \textbf{72.9} & \textbf{39.3} \\

  \hline
\end{tabular}
}

\vspace{0.1in}

\resizebox{\textwidth}{!} {
\begin{tabular}{|r|ccccccccccccc||c|}
  \hline
    & \textbf{kid} & \textbf{baby} & \textbf{tanktop} & \textbf{tshirt} & \textbf{casualjacket} & \textbf{mensuit} & \textbf{longskirt} & \textbf{shortskirt} & \textbf{smallshorts} & \textbf{lowcuttop} & \textbf{swimsuit} & \textbf{weddingdress} & \textbf{bermudashorts} & \textbf{mAP}  \\ \hline \hline
  \textbf{EPM} \cite{sharma13h} & 49.7 & 24.3 &37.7 & 61.6 & 40.0 & 57.1 & 44.8 & 39.0 & 46.8 & 61.3 & 32.2 & 64.2 & 43.7 & 58.7 \\
  \textbf{RAD} \cite{Joo13h} & 53.5 & 16.3 &37.0 & 67.1 & 42.6 & 64.8 & 42.0 & 30.1 & 49.6 & 66.0 & 46.7 & 62.1 & 42.0 & 59.3 \\
   \textbf{SM-SP} \cite{khan14g} & 55.9 & 18.3 &40.6 & 65.6 & 40.6 & 57.4 & 33.3 & 38.9 & 44.0 & 67.7 & 46.7 & 46.3 & 38.6 & 57.6 \\
   \textbf{D-EPM} \cite{Sharma15g} & 62.5 & \textbf{39.5} &48.4 & 75.1 & \textbf{63.5} & \textbf{75.9} & \textbf{67.3} & 52.6 & 56.6 & 84.6 & \textbf{67.8} & \textbf{79.7} & 53.1 & 69.6 \\
\hline\hline
    \textbf{This Paper} & \textbf{70.5} & 31.3 &\textbf{56.5} & \textbf{80.4} & 62.8 & 69.2 & 62.0 & \textbf{52.9} & \textbf{66.4} & \textbf{84.7} & 63.5 & 72.5 & \textbf{65.2} & \textbf{70.6} \\

  \hline
\end{tabular}
}
\caption{Comparison of our approach with the state-of-the-art on the 27
  Human Attributes (HAT-27) dataset. Our method, without using part-based information, achieves the best
  performance compared the state-of-the-art D-EPM method~\cite{Sharma15g} exploiting the part-based information.
}
\label{HAT27_soa1}
\end{table*}

\begin{figure*}[t!]
\centering
\subfigure[Class: crouching]{
   \includegraphics[scale =0.39] {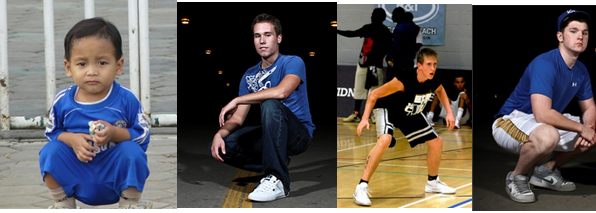}
 }
 \subfigure[Class: wedding dress ]{
   \includegraphics[scale =0.39] {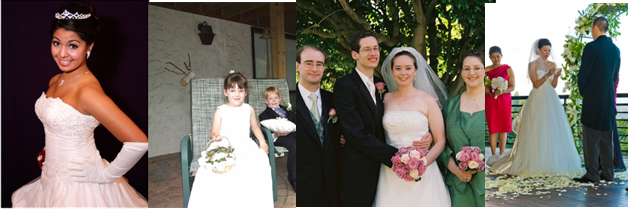}
 }
  \subfigure[Class: tank top]{
   \includegraphics[scale =0.39] {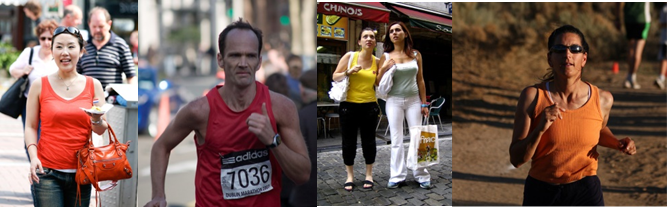}
 }
 \subfigure[Class: elderly]{
   \includegraphics[scale =0.39] {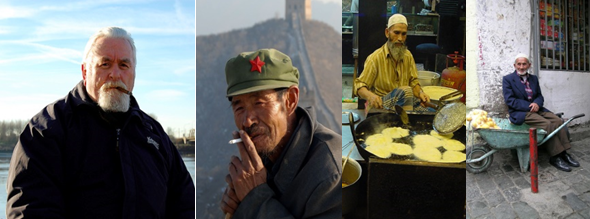}
 }
 \subfigure[Class: young]{
   \includegraphics[scale =0.39] {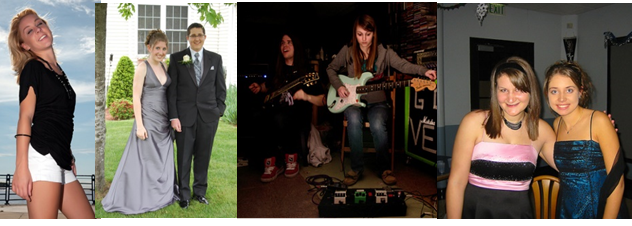}
 }
  \subfigure[Attribute category: baby]{
   \includegraphics[scale =0.39] {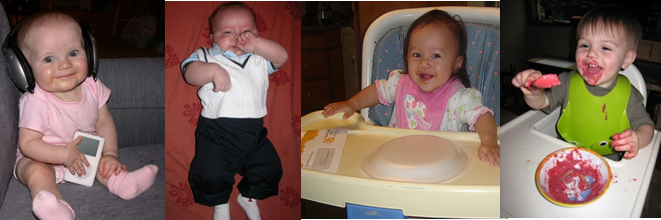}
 }
 \caption{Attribute classification performance of our approach on the
   HAT-27 dataset. We show top correct predictions of six attribute
   categories: `crouching', `wedding dress', `tank top', `elderly',
   `young' and `baby'.}
 \label{myfigure_hat_ex}
\end{figure*}

In Figure~\ref{fig:per_class_stanford} we compare the per-category
performance of our approach with two state-of-the-art approaches:
D-EPM \citep{Sharma15g} and FV-CNN \citep{Cimpoi15c}. Our scale coding
based approach achieves the best performance on 37 out of 40 action
categories on this dataset. A significant gain in performance is
achieved especially for drinking (+$14\%$), washing dishes (+$9\%$),
taking photos (+$9\%$), smoking (+$8\%$), and waving hands (+$8\%$)
action categories, all compared to the two state-of-the-art
methods. Table~\ref{stanford_rank_tab} shows example images from
pouring liquid, gardening, using computer and fishing categories. The
corresponding ranks are shown for the standard VGG-19 FC, FV-CNN and
our scale coding based approach. The number indicates
the absolute rank of corresponding image in the list of all test
images sorted by the probability for the corresponding class. A lower number implies higher
confidence in the action class label. We also show rank with respect to the number of false
positives appearing before the example test image in the
ranked list. Our approach obtains improved rank on these images compared to the two standard approaches.

\minisection{Human Attributes (HAT-27) dataset:} Finally,
Table~\ref{HAT27_soa1} shows a comparison of our scale coding based
approach with state-of-the-art methods on the Human Attributes
(HAT-27) dataset. The dataset contains 27 different human
attributes. The expanded part-based approach by Sharma et
al. \cite{sharma13h} yields an mAP of $58.7\%$, and semantic
pyramids \citep{khan14g}, combining body part information in a spatial
pyramid representation, an mAP of $57.6\%$. The approach
of \cite{Joo13h} is based on learning a rich appearance part based
dictionary and achieves an mAP of $59.3\%$. Deep FC features from the
VGG-19 network obtains a mAP score of $62.1\%$. The D-EPM
method \citep{Sharma15g} based on deep features and expanded part based
models achieves the best results among the existing methods with a mAP
of $69.6\%$. On this dataset, our scale coding based approach
outperforms the D-EPM method with a mAP score of $70.6\%$. Scale
coding yields the best classification performance on 15 out of 27
attribute categories compared to the state-of-the-art.

Figure~\ref{myfigure_hat_ex} illustrates the top four predictions of
six attribute categories from the HAT-27 dataset. These examples show
inter- and intra-class variations among different categories. The
variations in scale and pose of persons make the problem of attribute
classification challenging. Our scale coding based approach
consistently improves the performance on this dataset.

\subsubsection{Generality of Our Approach}
We have validated our approach on two challenging problems: human attribute and action classification. However, our scale coding approach is generic and is more broadly applicable to other recognition tasks. To validate the generality of our approach, we perform additional experiments on the popular MIT indoor scene 67 dataset \citep{Quattonicvpr09} for scene recognition task. The dataset contains 15620 images of 67 indoor scene classes. The training and test configurations are provided by the original authors, where each category has
around 80 images for training and 20 for testing. The performance is measured in terms of mean classification accuracy computed over all the categories in the dataset. Most existing methods \citep{Tsungiccv15,Weiiccv15,Kulkarnieccv16,Herranzcvpr16} report results using VGG16 model, pre-trained on either ImageNet or Places dataset. For fair comparison, we also validate our absolute scale coding approach using the VGG16 model and only compare with approaches pre-trained on ImageNet dataset.

Table~\ref{MIT67_soa} shows a comparison of our absolute scale coding based
approach with state-of-the-art methods on the MIT indoor scene 67 dataset. Among existing approaches, the work of \cite{Herranzcvpr16} also investigated multi-scale CNN architecture by training scale-specific networks on the ImageNet dataset, focusing on the CNN models. Several scale specific networks are combined by concatenating the FC7 features of all networks, yielding a mean accuracy score of $79.0\%$. Instead, our approach proposes multi-scale image representations by using a single pre-trained deep network and preserving scale information in the pooling method, obtaining a mean classification score of $81.9\%$. The results are further improved when combining the standard FC features with our scale coding approach. It is worth to mention that a higher recognition score of $86.0\%$ is obtained by \cite{Herranzcvpr16}, when combining scale-specific networks trained on both ImageNet and Places scene dataset. However, when using the same deep model architecture (VGG16) and only ImageNet dataset for network training, our results of $83.1\%$ are superior compared to $79.0\%$ obtained by the multi-scale scale-specific networks \citep{Herranzcvpr16}.

\begin{table}[t]
\centering \small\addtolength{\tabcolsep}{0.1pt}

\begin{tabular}{|l||c|}
\hline Method  &   Accuracy \\
\hline\hline DAG-CNN \cite{Songfaniccv15}   & 77.5 \\
\hline\hline Deep Spatial Pyramid \cite{Tsungiccv15}   & 78.3 \\
\hline\hline B-CNN \cite{Tsungiccv15}   & 79.0 \\
\hline\hline FV-CNN  \cite{Cimpoi15c}  & 79.2 \\
\hline\hline SPLeap  \cite{Kulkarnieccv16}  & 73.5 \\
\hline\hline Standard VGG16 \cite{Simonyan15j}   & 69.6 \\
\hline\hline Standard VGG16 FT  \cite{Herranzcvpr16}  & 76.4 \\
\hline\hline Multi-Scale Network \cite{Herranzcvpr16}   & 79.0 \\
\hline\hline Our Approach    & 81.9 \\
\hline\hline Our Approach + Standard VGG16   & \textbf{83.1} \\
\hline

\end{tabular}

\caption{
  Comparison of our approach with the state-of-the-art on the MIT indoor scene 67 dataset. Our method achieves superior performance compared to existing approaches based on the same VGG16 model.
}
\label{MIT67_soa}
\end{table}

\section{Conclusions}
\label{conclusions}
In this paper we investigated the problem of encoding multi-scale
information for still images in the context of human attribute and
action recognition. Most state-of-the-art approaches based on the BOW
framework compute local descriptors at multiple scales. However,
multi-scale information is not explicitly encoded as all the features
from different scales are pooled into a single scale-invariant
histogram. In the context of human attribute and action recognition,
we demonstrate that both absolute and relative scale information can
be encoded in final image representations and that relaxing the
traditional scale invariance commonly employed in image classification
can lead to significant gains in recognition performance.

We proposed two alternative scale coding approaches that explicitly
encode scale information in the final image representation. The
absolute scale of local features is encoded by constructing separate
representations for small, medium and large features, while the relative
scale of the local features is encoded with respect to the size of the
bounding box corresponding to the person instance in human action or
attribute recognition problems. In both cases, the final image
representation is obtained by concatenating the small, medium and
large scale representations.

Comprehensive experiments on five datasets demonstrate the
effectiveness of our proposed approach. The results clearly
demonstrate that our scale coding strategies outperform both the
scale-invariant bag of deep features and the standard deep features
extracted from the same network. An interesting future direction is
the investigation of scale coding strategies for object detection and
fine-grained object localization. We believe that our scale coding
schemes could be very effective for representing candidate regions in
object detection techniques based on bottom-up proposal of likely
object regions.

\section*{Acknowledgements}

\noindent This work has been funded by the projects TIN2013-41751 and of the Spanish Ministry of Science, the Catalan project 2014 SGR 221, the CHISTERA project PCIN-2015-251, SSF through a grant for the project SymbiCloud, VR (EMC${}^2$), VR starting grant (2016-05543), through the Strategic Area for ICT research ELLIIT, the grant 251170 of the Academy of Finland. The calculations were performed using computer resources within the Aalto University School of Science ``Science-IT'' project. We also acknowledge the support from Nvidia and the NSC.

\bibliographystyle{spbasic}      
\bibliography{Gender_Action_IEEE}

\begin{thebibliography}{72}
\providecommand{\natexlab}[1]{#1}
\providecommand{\url}[1]{{#1}}
\providecommand{\urlprefix}{URL }
\expandafter\ifx\csname urlstyle\endcsname\relax
  \providecommand{\doi}[1]{DOI~\discretionary{}{}{}#1}\else
  \providecommand{\doi}{DOI~\discretionary{}{}{}\begingroup
  \urlstyle{rm}\Url}\fi
\providecommand{\eprint}[2][]{\url{#2}}

\bibitem[{Azizpour et~al(2014)Azizpour, Sullivan, and Carlsson}]{Azizpour14h}
Azizpour H, Sullivan J, Carlsson S (2014) Cnn features off-the-shelf: An
  astounding baseline for recognition. In: CVPRW, pp 512--519

\bibitem[{Bosch et~al(2008)Bosch, Zisserman, and Munoz}]{Bosch08g}
Bosch A, Zisserman A, Munoz X (2008) Scene classification using a hybrid
  generative/discriminative approach. PAMI 30(4):712--727

\bibitem[{Bourdev et~al(2011)Bourdev, Maji, and Malik}]{Bourdev11d}
Bourdev L, Maji S, Malik J (2011) Describing people: A poselet-based approach
  to attribute classification. In: ICCV, pp 1543--1550

\bibitem[{Chatfield et~al(2011)Chatfield, Lempitsky, Vedaldi, and
  Zisserman}]{chatfield2011devil}
Chatfield K, Lempitsky V, Vedaldi A, Zisserman A (2011) The devil is in the
  details: an evaluation of recent feature encoding methods. In: BMVC

\bibitem[{Chatfield et~al(2014)Chatfield, Simonyan, Vedaldi, and
  Zisserman}]{Chatfield14h}
Chatfield K, Simonyan K, Vedaldi A, Zisserman A (2014) Return of the devil in
  the details: Delving deep into convolutional nets. In: BMVC

\bibitem[{Cimpoi et~al(2015)Cimpoi, Maji, and Vedaldi}]{Cimpoi15c}
Cimpoi M, Maji S, Vedaldi A (2015) Deep filter banks for texture recognition
  and segmentation. In: CVPR, pp 3828--3836

\bibitem[{Dalal and Triggs(2005)}]{Dalal05}
Dalal N, Triggs B (2005) Histograms of oriented gradients for human detection.
  In: CVPR, pp 886--893

\bibitem[{Delaitre et~al(2010)Delaitre, Laptev, and Sivic}]{Laptev10}
Delaitre V, Laptev I, Sivic J (2010) Recognizing human actions in still images:
  a study of bag-of-features and part-based representations. In: BMVC

\bibitem[{Delaitre et~al(2011)Delaitre, Sivic, and Laptev}]{Laptev11j}
Delaitre V, Sivic J, Laptev I (2011) Learning person-object interactions for
  action recognition in still images. In: NIPS, pp 1503--1511

\bibitem[{Everingham et~al(2010)Everingham, Gool, Williams, Winn, and
  Zisserman}]{everingham10b}
Everingham M, Gool LJV, Williams CKI, Winn JM, Zisserman A (2010) The pascal
  visual object classes (voc) challenge. IJCV 88(2):303--338

\bibitem[{van Gemert et~al(2010)van Gemert, Veenman, Smeulders, and
  Geusebroek}]{Gemert10g}
van Gemert J, Veenman C, Smeulders A, Geusebroek JM (2010) Visual word
  ambiguity. PAMI 32(7):1271--1283

\bibitem[{Girshick et~al(2014)Girshick, Donahue, Darrell, and
  Malik}]{Girshick14h}
Girshick R, Donahue J, Darrell T, Malik J (2014) Rich feature hierarchies for
  accurate object detection and semantic segmentation. In: CVPR, pp 580--587

\bibitem[{Gkioxari et~al(2014)Gkioxari, Hariharan, Girshick, and
  Malik}]{Gkioxari14g}
Gkioxari G, Hariharan B, Girshick R, Malik J (2014) R-cnns for pose estimation
  and action detection. arXiv preprint arXiv:14065212

\bibitem[{Gong et~al(2014)Gong, Wang, Guo, and Lazebnik}]{Gong14c}
Gong Y, Wang L, Guo R, Lazebnik S (2014) Multi-scale orderless pooling of deep
  convolutional activation features. In: ECCV, pp 392--407

\bibitem[{Guo and Lai(2014)}]{Guo14c}
Guo G, Lai A (2014) A survey on still image based human action recognition. PR
  47(10):3343--3361

\bibitem[{Harzallah et~al(2009)Harzallah, Jurie, and Schmid}]{HJS09}
Harzallah H, Jurie F, Schmid C (2009) Combining efficient object localization
  and image classification. In: ICCV

\bibitem[{Herranz et~al(2016)Herranz, Jiang, and Li}]{Herranzcvpr16}
Herranz L, Jiang S, Li X (2016) Scene recognition with cnns: Objects, scales
  and dataset bias. In: CVPR

\bibitem[{Hoai(2014)}]{Hoai14k}
Hoai M (2014) Regularized max pooling for image categorization. In: BMVC

\bibitem[{Hoai et~al(2014)Hoai, Ladicky, and Zisserman}]{Hoai14h}
Hoai M, Ladicky L, Zisserman A (2014) Action recognition from weak alignment of
  body parts. In: BMVC

\bibitem[{Jegou et~al(2010)Jegou, Douze, and Schmid}]{Jegou10b}
Jegou H, Douze M, Schmid C (2010) Improving bag-of-features for large scale
  image search. IJCV 87(3):316--336

\bibitem[{J{\'e}gou et~al(2010)J{\'e}gou, Douze, Schmid, and
  P{\'e}rez}]{jegou2010aggregating}
J{\'e}gou H, Douze M, Schmid C, P{\'e}rez P (2010) Aggregating local
  descriptors into a compact image representation. In: CVPR, pp 3304--3311

\bibitem[{Joo et~al(2013)Joo, Wang, and Zhu}]{Joo13h}
Joo J, Wang S, Zhu SC (2013) Human attribute recognition by rich appearance
  dictionary. In: ICCV, pp 721--728

\bibitem[{Khan et~al(2012)Khan, van~de Weijer, and Vanrell}]{fahad12b}
Khan FS, van~de Weijer J, Vanrell M (2012) Modulating shape features by color
  attention for object recognition. IJCV 98(1):49--64

\bibitem[{Khan et~al(2013)Khan, Anwer, van~de Weijer, Bagdanov, Lopez, and
  Felsberg}]{fahad13b}
Khan FS, Anwer RM, van~de Weijer J, Bagdanov A, Lopez A, Felsberg M (2013)
  Coloring action recognition in still images. IJCV 105(3):205--221

\bibitem[{Khan et~al(2014{\natexlab{a}})Khan, van~de Weijer, Anwer, Felsberg,
  and Gatta}]{khan14g}
Khan FS, van~de Weijer J, Anwer RM, Felsberg M, Gatta C (2014{\natexlab{a}})
  Semantic pyramids for gender and action recognition. TIP 23(8):3633--3645

\bibitem[{Khan et~al(2014{\natexlab{b}})Khan, van~de Weijer, Bagdanov, and
  Felsberg}]{fahad14h}
Khan FS, van~de Weijer J, Bagdanov A, Felsberg M (2014{\natexlab{b}}) Scale
  coding bag-of-words for action recognition. In: ICPR, pp 1514--1519

\bibitem[{Khan et~al(2015)Khan, Xu, van~de Weijer, Bagdanov, Anwer, and
  Lopez}]{Khan15k}
Khan FS, Xu J, van~de Weijer J, Bagdanov A, Anwer RM, Lopez A (2015)
  Recognizing actions through action-specific person detection. TIP
  24(11):4422--4432

\bibitem[{Koenderink(1984)}]{Koenderink84g}
Koenderink J (1984) The structure of images. Biological cybernetics
  50(5):363--370

\bibitem[{Koskela and Laaksonen(2014)}]{Koskela14h}
Koskela M, Laaksonen J (2014) Convolutional network features for scene
  recognition. In: ACM Multimedia, pp 1169--1172

\bibitem[{Kulkarni et~al(2016)Kulkarni, Jurie, Zepeda, Perez, and
  Chevallie}]{Kulkarnieccv16}
Kulkarni P, Jurie F, Zepeda J, Perez P, Chevallie L (2016) Spleap: Soft pooling
  of learned parts for image classification. In: ECCV

\bibitem[{Lazebnik et~al(2006)Lazebnik, Schmid, and Ponce}]{Lazebnik06}
Lazebnik S, Schmid C, Ponce J (2006) Beyond bags of features: Spatial pyramid
  matching for recognizing natural scene categories. In: CVPR, pp 2169--2178

\bibitem[{LeCun et~al(1989)LeCun, Boser, Denker, Henderson, Howard, Hubbard,
  and Jackel}]{LeCun89h}
LeCun Y, Boser B, Denker J, Henderson D, Howard R, Hubbard W, Jackel L (1989)
  Handwritten digit recognition with a back-propagation network. In: NIPS, pp
  396--404

\bibitem[{Liang et~al(2014)Liang, Wang, Huang, and Lin}]{Liang14h}
Liang Z, Wang X, Huang R, Lin L (2014) An expressive deep model for human
  action parsing from a single image. In: ICME, pp 1--6

\bibitem[{Lim et~al(2015)Lim, Vats, and Chan}]{Hong15c}
Lim CH, Vats E, Chan CS (2015) Fuzzy human motion analysis: A review. PR
  48(5):1773--1796

\bibitem[{Lin et~al(2015)Lin, RoyChowdhury, and Maji}]{Tsungiccv15}
Lin TY, RoyChowdhury A, Maji S (2015) Bilinear cnn models for fine-grained
  visual recognition. In: ICCV

\bibitem[{Liu et~al(2015)Liu, Shen, and van~den Hengel}]{Lingqiao15c}
Liu L, Shen C, van~den Hengel A (2015) The treasure beneath convolutional
  layers: Cross-convolutional-layer pooling for image classification. In: CVPR,
  pp 4749--4757

\bibitem[{Lowe(2004)}]{Lowe04}
Lowe D (2004) Distinctive image features from scale-invariant points. IJCV
  60(2):91--110

\bibitem[{Maji et~al(2011)Maji, Bourdev, and Malik}]{Maji11f}
Maji S, Bourdev LD, Malik J (2011) Action recognition from a distributed
  representation of pose and appearance. In: CVPR, pp 3177--3184

\bibitem[{Maragos(1989)}]{Maragos89g}
Maragos P (1989) Pattern spectrum and multiscale shape representation. PAMI
  11(7):701--716

\bibitem[{Mettes et~al(2016)Mettes, van Gemer, and Snoek}]{Mettes15g}
Mettes P, van Gemer J, Snoek C (2016) No spare parts: Sharing part detectors
  for image categorization. CVIU 152:131--141

\bibitem[{Mikolajczyk and Schmid(2004{\natexlab{a}})}]{Mikolajczyk04gg}
Mikolajczyk K, Schmid C (2004{\natexlab{a}}) Scale and affine invariant
  interest point detectors. IJCV 60(1):63--86

\bibitem[{Mikolajczyk and Schmid(2004{\natexlab{b}})}]{Mikolajczyk04a}
Mikolajczyk K, Schmid C (2004{\natexlab{b}}) Scale and affine invariant
  interest point detectors. IJCV 60(1):63--86

\bibitem[{Nowak et~al(2006)Nowak, Jurie, and Triggs}]{Jurie06}
Nowak E, Jurie F, Triggs B (2006) Sampling strategies for bag-of-features image
  classification. In: ECCV, pp 490--503

\bibitem[{Oquab et~al(2014)Oquab, Bottou, Laptev, and Sivic}]{Oquab14k}
Oquab M, Bottou L, Laptev I, Sivic J (2014) Learning and transferring mid-level
  image representations using convolutional neural networks. In: CVPR, pp
  1717--1724

\bibitem[{Perronnin and Dance(2007)}]{perronnin2007fisher}
Perronnin F, Dance C (2007) Fisher kernels on visual vocabularies for image
  categorization. In: CVPR, pp 1--8

\bibitem[{Perronnin et~al(2010)Perronnin, Sanchez, and Mensink}]{Perronnin10c}
Perronnin F, Sanchez J, Mensink T (2010) Improving the fisher kernel for
  large-scale image classification. In: ECCV, pp 143--156

\bibitem[{Prest et~al(2012)Prest, Schmid, and Ferrari}]{Cordelia12c}
Prest A, Schmid C, Ferrari V (2012) Weakly supervised learning of interactions
  between humans and objects. PAMI 34(3):601--614

\bibitem[{Quattoni and Torralba(2009)}]{Quattonicvpr09}
Quattoni A, Torralba A (2009) Recognizing indoor scenes. In: CVPR

\bibitem[{Rojas et~al(2010)Rojas, Khan, van~de Weijer, and Gevers}]{Rojas10j}
Rojas D, Khan FS, van~de Weijer J, Gevers T (2010) The impact of color on
  bag-of-words based object recognition. In: ICPR, pp 1549--1553

\bibitem[{Russakovsky et~al(2014)Russakovsky, Deng, Su, Krause, Satheesh, Ma,
  Huang, Karpathy, Khosla, Bernstein et~al}]{ImageNet:2014}
Russakovsky O, Deng J, Su H, Krause J, Satheesh S, Ma S, Huang Z, Karpathy A,
  Khosla A, Bernstein M, et~al (2014) Imagenet large scale visual recognition
  challenge. arXiv preprint arXiv:14090575

\bibitem[{S{\'a}nchez et~al(2013)S{\'a}nchez, Perronnin, Mensink, and
  Verbeek}]{sanchez2013image}
S{\'a}nchez J, Perronnin F, Mensink T, Verbeek J (2013) Image classification
  with the fisher vector: Theory and practice. International journal of
  computer vision 105(3):222--245

\bibitem[{van~de Sande et~al(2010)van~de Sande, Gevers, and Snoek}]{gevers10}
van~de Sande K, Gevers T, Snoek CGM (2010) Evaluating color descriptors for
  object and scene recognition. PAMI 32(9):1582--1596

\bibitem[{Shabani et~al(2013)Shabani, Zelek, and Clausi}]{Shabani13c}
Shabani AH, Zelek J, Clausi D (2013) Multiple scale-specific representations
  for improved human action recognition. PRL 34(15):1771--1779

\bibitem[{Shapovalova et~al(2011)Shapovalova, Gong, Pedersoli, Roca, and
  Gonzalez}]{Pedersoli11}
Shapovalova N, Gong W, Pedersoli M, Roca FX, Gonzalez J (2011) On importance of
  interactions and context in human action recognition. In: IbPRIA, pp 58--66

\bibitem[{Sharma and Jurie(2011)}]{SharmaHAT27:2011}
Sharma G, Jurie F (2011) Learning discriminative spatial representation for
  image classification. In: BMVC

\bibitem[{Sharma et~al(2012)Sharma, Jurie, and Schmid}]{schmid12}
Sharma G, Jurie F, Schmid C (2012) Discriminative spatial saliency for image
  classification. In: CVPR, pp 3506--3513

\bibitem[{Sharma et~al(2013)Sharma, Jurie, and Schmid}]{sharma13h}
Sharma G, Jurie F, Schmid C (2013) Expanded parts model for human attribute and
  action recognition in still images. In: CVPR, pp 652--659

\bibitem[{Sharma et~al(2015)Sharma, Jurie, and Schmid}]{Sharma15g}
Sharma G, Jurie F, Schmid C (2015) Expanded parts model for semantic
  description of humans in still images. arXiv preprint, arXiv:150904186

\bibitem[{Sicre and Jurie(2015)}]{Sicre15c}
Sicre R, Jurie F (2015) Discriminative part model for visual recognition. CVIU
  141:28--37

\bibitem[{Simonyan and Zisserman(2015)}]{Simonyan15j}
Simonyan K, Zisserman A (2015) Very deep convolutional networks for large-scale
  image recognition. In: ICLR

\bibitem[{Vedaldi and Fulkerson(2010)}]{Vedaldi10b}
Vedaldi A, Fulkerson B (2010) Vlfeat: an open and portable library of computer
  vision algorithms. In: ACM MM, pp 1469--1472

\bibitem[{Vedaldi et~al(2009)Vedaldi, Gulshan, Varma, and
  Zisserman}]{vedaldi09h}
Vedaldi A, Gulshan V, Varma M, Zisserman A (2009) Multiple kernels for object
  detection. In: ICCV, pp 606--613

\bibitem[{Wei et~al(2015)Wei, Gao, and Wu}]{Weiiccv15}
Wei XS, Gao BB, Wu J (2015) Deep spatial pyramid ensemble for cultural event
  recognition. In: ICCV Workshop

\bibitem[{van~de Weijer et~al(2009)van~de Weijer, Schmid, Verbeek, and
  Larlus}]{Weijer09a}
van~de Weijer J, Schmid C, Verbeek JJ, Larlus D (2009) Learning color names for
  real-world applications. TIP 18(7):1512--1524

\bibitem[{Witkin(1984)}]{Witkin84}
Witkin A (1984) Scale-space filtering: A new approach to multi-scale
  description. In: ICASSP

\bibitem[{Yang and Ramanan(2015)}]{Songfaniccv15}
Yang S, Ramanan D (2015) Multi-scale recognition with dag-cnns. In: ICCV

\bibitem[{Yao et~al(2011)Yao, Jiang, Khosla, Lin, Guibas, and Li}]{Yao11f}
Yao B, Jiang X, Khosla A, Lin AL, Guibas LJ, Li FF (2011) Human action
  recognition by learning bases of action attributes and parts. In: ICCV, pp
  1331--1338

\bibitem[{Zhang et~al(2014)Zhang, Paluri, Ranzato, Darrell, and
  Bourdev}]{Zhang14h}
Zhang N, Paluri M, Ranzato M, Darrell T, Bourdev L (2014) Panda: Pose aligned
  networks for deep attribute modeling. In: CVPR, pp 1637--1644

\bibitem[{Zhou et~al(2014)Zhou, Lapedriza, Xiao, Torralba, and Oliva}]{Zhou14k}
Zhou B, Lapedriza A, Xiao J, Torralba A, Oliva A (2014) Learning deep features
  for scene recognition using places database. In: NIPS, pp 487--495

\bibitem[{Zhou et~al(2010)Zhou, Yu, Zhang, and Huang}]{Zhou10h}
Zhou X, Yu K, Zhang T, Huang T (2010) Image classification using super-vector
  coding of local image descriptors. In: ECCV, pp 141--154

\bibitem[{Zhu et~al(2012)Zhu, Li, Li, Yang, and Tsien}]{Zhu13k}
Zhu X, Li M, Li X, Yang Z, Tsien J (2012) Robust action recognition using
  multi-scale spatial-temporal concatenations of local features as natural
  action structures. PLoS One 7(10)

\bibitem[{Ziaeefard and Bergevin(2015)}]{Ziaeefard15c}
Ziaeefard M, Bergevin R (2015) Semantic human activity recognition: A
  literature review. PR 48(8):2329--2345

\end{thebibliography}

\end{document}